\documentclass[letterpaper]{article} 
\usepackage{aaai24}  
\usepackage{times}  
\usepackage{helvet}  
\usepackage{courier}  
\usepackage[hyphens]{url}  
\usepackage{graphicx} 
\usepackage{natbib}  
\usepackage{caption} 
\usepackage{algorithm}
\usepackage{algorithmic}

\usepackage{newfloat}
\usepackage{listings}
\DeclareCaptionStyle{ruled}{labelfont=normalfont,labelsep=colon,strut=off} 
\floatstyle{ruled}
\newfloat{listing}{tb}{lst}{}
\floatname{listing}{Listing}
\usepackage{amsmath,amsfonts,bm}

\DeclareMathOperator*{\argmax}{arg\,max} 
\newcommand{\phiminus}{\Phi^{-1}}
\newcommand{\pabar}{\underline{p_A}(\sigma)}

\DeclareMathOperator{\dom}{dom}

\def\eqref#1{equation~\ref{#1}}

\def\1{\bm{1}}

\DeclareMathAlphabet{\mathsfit}{\encodingdefault}{\sfdefault}{m}{sl}
\SetMathAlphabet{\mathsfit}{bold}{\encodingdefault}{\sfdefault}{bx}{n}

\usepackage{url}            
\usepackage{booktabs}       
\usepackage{amsfonts}       
\usepackage{xcolor}         
\usepackage{amsmath}
\usepackage{amsthm}
\usepackage{amssymb}
\usepackage{subfig}

\newtheorem{theorem}{Theorem}
\newtheorem{lemma}{Lemma}
\newtheorem{definition}{Definition}
\newtheorem{proposition}{Proposition}

\title{Towards Large Certified Radius in Randomized Smoothing using\\ Quasiconcave Optimization}

\author {
    Bo-Han~Kung,
    Shang-Tse~Chen
}
\affiliations {
    National Taiwan University\\
    \texttt{\{d10922019, stchen\}@csie.ntu.edu.tw}
}

\begin{document}

\newcommand{\fix}{\marginpar{FIX}}
\newcommand{\new}{\marginpar{NEW}}
\newcommand{\cohen}{\textsc{Cohen} }
\newcommand{\terma}{$\mathbf{A}$ }
\newcommand{\termb}{$\mathbf{B}$ }

\maketitle

\begin{abstract}
Randomized smoothing is currently the state-of-the-art method that provides certified robustness for deep neural networks. However, due to its excessively conservative nature, this method of incomplete verification often cannot achieve an adequate certified radius on real-world datasets. One way to obtain a larger certified radius is to use an input-specific algorithm instead of using a fixed Gaussian filter for all data points. Several methods based on this idea have been proposed, but they either suffer from high computational costs or gain marginal improvement in certified radius. In this work, we show that by exploiting the quasiconvex problem structure, we can find the optimal certified radii for most data points with slight computational overhead. This observation leads to an efficient and effective input-specific randomized smoothing algorithm. We conduct extensive experiments and empirical analysis on CIFAR-10 and ImageNet. The results show that the proposed method significantly enhances the certified radii with low computational overhead. 
\end{abstract}

\section{Introduction}
Although deep learning has achieved tremendous success in various fields~\cite{wang2022yolov7,zhai2022scaling}, it is known to be vulnerable to adversarial attacks \cite{szegedy2013intriguing}.
This kind of attack crafts an imperceptible perturbation on images \cite{goodfellow2014explaining} or voices \cite{carlini2018audio} to make the AI system predict incorrectly.
Many adversarial defense methods have been proposed to defend against adversarial attacks.
Adversarial defenses can be categorized into  empirical defenses and theoretical defenses.
Common empirical defenses include adversarial training \cite{madry2017towards,shafahi2019adversarial,wong2020fast} and preprocessing-based methods~\cite{defensegan,das2018shield}.
Though effective, empirical defenses cannot guarantee robustness.

Different from empirical defenses, theoretical defenses provide robustness verification with mathematical guarantees \cite{li2023sok}. 
The verification must be \textit{sound}, thereby preventing false positives. However, while it has the potential to be \textit{complete}, ensuring no false negatives, it may also be \textit{incomplete}.
One the other hand, theoretical defenses can be categorized into \textit{deterministic} verification and \textit{probabilistic} verification.
Deterministic verification methods, such as mixed-integer programming (MIP) \cite{tjeng2018evaluating}, interval bound propagation (IBP)\cite{ehlers2017formal,weng2018towards,gowal2018effectiveness,mueller2022certified} and Lipschitz-bounded networks \cite{singla2021skew,singla2021improved,xu2022lot}, offer theoretical robustness guarantees for a DNN model. The verification in these cases is deterministic.
In contrast, probabilistic verification methods, such as randomized smoothing \cite{cohen2019certified,lecuyer2019certified,yang2020randomized}, provide a provable defense that ensures that there are no adversarial examples within a specific ball with a radius $R$. Notably, this guarantee is accompanied by a degree of randomness that is independent of input images.

Among those methods, only randomized smoothing (RS) can scale to state-of-the-art deep neural networks and real-world datasets. 
Randomized smoothing first builds a smoothed classifier for a given data point via a Gaussian filter and Monte Carlo sampling, and then it estimates a confidence lower bound for the highest-probability class. 
Next, it determines a certified radius for the class and promise that there is no adversarial example within this radius.

\begin{figure*}
    \centering
    \includegraphics[width=.9\textwidth]{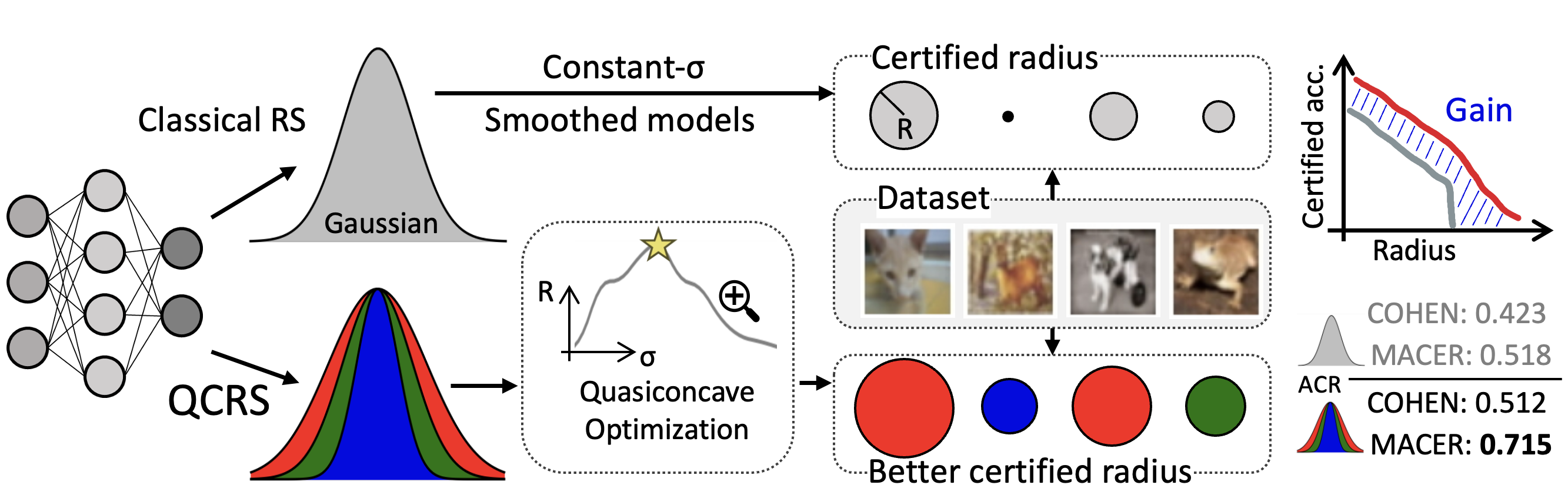}
    \vspace{-0.3cm}
    \caption{Overview of the proposed QCRS algorithm. QCRS finds the better sigma value for each smoothed classifier using quasiconcave optimization. Thus, it provides better certified radii than the classical randomized smoothing. In this paper, we discuss quasiconcavity on the certified radius w.r.t. $\sigma$. i.e., $R(\sigma)$.}\label{fig:teaser}
    \vspace{-0.3cm}
\end{figure*}

Although randomized smoothing is effective, it suffers from two main disadvantages.
First, randomized smoothing applies the same constant-variance Gaussian filter to every data point when constructing a smoothed classifier.
This makes the certified radius dramatically underestimated. 
Second, randomized smoothing adopts a confidence lower bound (Clopper-Pearson lower bound) to estimate the highest-probability class, which also limits the certified radius.
As a result, when evaluating radius-accuracy curve, a truncation fall often occurs (see the gray curve in the upper right of Fig.~\ref{fig:teaser}).
This is called \textit{truncation effect} or waterfall effect \cite{sukenik2021intriguing}, which shows the conservation aspect in randomized smoothing.
Other issues such as fairness \cite{mohapatra2021hidden}, dimension \cite{kumar2020curse}, and time-efficiency \cite{chen2022input} also limit its application.

To alleviate truncation effect and improve the certified radii, a more precise workflow is necessary.
Prior work \cite{chen2021insta,alfarra2020data} proposed input-specific methods that can assign different Gaussian filters to different data points.
Those methods attempt to optimize the radius by finding the optimal variance $\sigma^2$ of the Gaussian filter.
In this work, we first delve into randomized smoothing and discover a useful property called quasiconcavity for the sigma-radius curve.
Next, we define a probabilistic quasiconcavity assumption and then develop a novel algorithm called \textbf{Quasiconcavity-based Randomized Smoothing (QCRS)} that optimizes certified radii with respect to sigma.
The overview of QCRS is illustrated in Fig.~\ref{fig:teaser}.
QCRS significantly improves the certified radii with little computational overhead compared to existing methods \cite{chen2021insta,alfarra2020data}.
The proposed QCRS enjoys the advantages of both performance and time-efficiency.
The main technical contributions are summarized as follows:
\begin{itemize}
    \item We discover and empirically demonstrate that the sigma-radius curves are quasiconcave for most data points. In our experiments, approximately $99\%$ of the data points satisfy our proposed quasiconcavity condition.
    \item Based on the observed quasiconcavity property, we propose a novel and efficient input-specific algorithm called QCRS. This algorithm aims to enhance certified radii and alleviate the truncation effect in randomized smoothing.
    \item Through extensive experiments, we demonstrate the effectiveness of our proposed QCRS method on the CIFAR-10 and ImageNet datasets. Furthermore, by combining QCRS with a training-based method, we achieve state-of-the-art certified radii.
\end{itemize}

\section{Related Work}
Randomized smoothing utilizes a spatial low-pass Gaussian filter to construct a smoothed model~\cite{cohen2019certified}.
Based on the Neyman-Pearson lemma, this smoothed model can provide a provable radius $R$ to guarantee robustness for large-scale datasets.
To improve randomized smoothing, some works \cite{yang2020randomized,zhang2020black,levine2021improved} proposed general methods using different smoothing distribution for different $\ell_{p}$ balls, while others tried to provide a better and tighter certification \cite{kumar2020certifying,levine2020tight}.

\noindent \textbf{Improving RS during training phase}. 
To further enlarge the radius $R$, some works used training-based method \cite{salman2019provably,zhai2019macer,jeong2021smoothmix,anderson2022certified}.
These models were specifically designed for randomized smoothing.
For example, MACER \cite{zhai2019macer} made the computation of certified radius differentiable and add it to the standard cross-entropy loss. 
Thus, the average certified radius of MACER outperforms the Gaussian-augmentation model that was used by the original randomized smoothing.

\noindent \textbf{Improving RS during inference phase}.
Different from training-based method, some works utilized different smoothing methods to enhance the certified region.
Chen \textit{et al.,} \cite{chen2021insta} proposed a multiple-start search algorithm to find the best parameter for building smoothed classifiers.
Alfarra \textit{et al.,} \cite{alfarra2020data} adopted a memory-based approach to optimize the Gaussian filter of each input data.
Chen \textit{et al.,} \cite{chen2022input} proposed an input-specific sampling acceleration method to control the sampling number and provides fast and effective certification.
Li \textit{et al.,} \cite{li2022double} proposed double sampling randomized smoothing that utilizes additional smoothing information for tighter certification.
These inference-time methods are the most relevant to our work. See Section~\ref{sec:obser} for more detailed description on these methods.

\section{Preliminaries}
Let $x \in \mathbb{R}^d$ be a data point, where $d$ is the input dimension.
$\mathcal{C}=\{1,2,...,c\}$ is the set of classes.
$F:\mathbb{R}^d \rightarrow \mathbb{R}^c$ is a general predictor such as neural networks.
We define the base classifier as
\begin{equation}
    f(x)=e_{\xi};\quad \xi=\argmax_j F_j(x),
\end{equation}
where $e_j$ denotes a one-hot vector where the $j^{th}$ component is 1 and all the other components are 0.
The smoothed classifier \cite{cohen2019certified} $g:\mathbb{R}^d \rightarrow \mathcal{C}$ is defined as 
\begin{equation}
    g(x)=\argmax_{c\in \mathcal{C}} Pr[f(x+\epsilon)=e_c], \quad \epsilon \sim \mathcal{N}(0,\sigma^2 I),
\end{equation}
where $\mathcal{N}$ is Gaussian distribution and $\epsilon$ is a noise vector sampled from $\mathcal{N}$.
Cohen \textit{et al.,} (\textsc{Cohen}) proposed a provable method to calculate the certifiable robust radius as follows:
\begin{equation}\label{eq:cohen}
\begin{aligned}
        R = \frac{\sigma}{2} \cdot [\phiminus(\underline{p_A}) - \phiminus(\overline{p_B})], 
\end{aligned}
\end{equation}
\[
p_{A}=Pr[f(x+\epsilon)=e_A], \text{ and }  p_{B}=Pr[f(x+\epsilon)=e_B],
\]
where $A$ is the highest-probability class of the smoothed classifier, and $B$ is the runner-up class.
$\underline{p_A}$ and $\overline{p_B}$ are the Clopper-Pearson lower/upper bound of $p_A$ and $p_B$, which can be estimated by Monte Carlo (MC) sampling with a confidence level $1-\alpha$.
$R$ indicates the certified radius. Any data point inside this radius would be predicted as class $A$ by the smoothed classifier. That is, $g(x+\delta)=g(x)$ for all $||\delta||_2 \leq R$.
In practice, \cohen replaces $\overline{p_B}$ with $1-\underline{p_A}$, so \eqref{eq:cohen} usually is reformulated as $R = \sigma \cdot \phiminus(\underline{p_A})$. 
If $\underline{p_A} < 0.5$, it indicates that there is no certified radius in this data point according to \textsc{Cohen}.

The smoothed classifier $g$ is constructed from the base classifier $f$ by introducing perturbations $\epsilon$ to $x$.
Therefore, the smoothed classifier $g$ can be regarded as a spatial smoothing measure of the original base classifier $f$ using a Gaussian kernel $\mathcal{G}$, i.e., $g = f \star \mathcal{G}$, where $\star$ is the convolution operator.
From the signal-processing perspective, $g$ is the Weierstrass transformation of $f$.
That is, the smoothed classifier $g$ is constructed by applying a low-pass filter $\mathcal{G}$ on $f$.
Randomized smoothing constructs smoothed classifier to provide certifiable robustness guarantee.

\section{QCRS Methodology}
\subsection{Observation and Motivation}\label{sec:obser}
As mentioned earlier, several existing methods attempt to address the truncation effect.
Some focus on training the base model to enlarge certified radii, while others use a different Gaussian kernel $\mathcal{G}(\sigma)$ for each image to construct $g$. We follow the later approach and propose an input-specific algorithm that finds the optimal  $\mathcal{G}$ for most data points.
Intuitively, for a data point $x$ of class $c_A$, if most neighboring points belong to the same class $c_A$, we can use $\mathcal{G}$ with a larger variance to convolute the data space of $f$.
In contrast, if the neighborhood is full of different class samples, $\mathcal{G}$ needs a small variance to prevent misclassification. 
The proposed method draws inspiration from this concept and aims to optimize the selection of $\mathcal{G}(\sigma)$. 
Below, we will discuss some input-specific search algorithms that have been utilized in prior works.
These algorithms contribute to the development of our approach.

DDRS \cite{alfarra2020data} assumes that sigma-radius curves, $R(\sigma) = \sigma \cdot \phiminus(\pabar)$, are concave and use gradient-based convex optimization along with some relaxation and approximation to find the optimal $\sigma$ value. 
However, in our experiments, almost all sigma-radius curves are not concave.
We select $200$ images from CIFAR-10 dataset and compute the certified radii with respect to $\sigma$ for each image. 
Among these $200$ images, as least $189$ images do not satisfy concavity.
Thus, the gradient-based convex optimization method may not work well in this task.
Instead of depending on the assumption of concavity, Insta-RS \cite{chen2021insta} uses a multi-start searching algorithm to optimize $\sigma$. However, the multi-start procedure incurs high computational overhead. 
In our work, we observe an intriguing quasiconcave property on the sigma-radius curves, which helps us develop an effective and efficient algorithm to optimize sigma.
In order to assess the generality of the quasiconcave property, we conducted evaluations on two datasets and four distinct models.
Table~\ref{tab:generalityOfQC} illustrates the generality of concavity and quasiconcavity of the sigma-radius curves.
The table demonstrates that quasiconcavity is significantly more prevalent than concavity in real-world datasets.
We will delve deeper into this table in Subsection~\ref{sec:design}.

\begin{table}[tp]
    \centering
    \scalebox{0.9}{
    \begin{tabular}{c|cccc}
    \toprule
         Dataset & \multicolumn{2}{c}{CIFAR-10}& \multicolumn{2}{c}{ImageNet}\\
         Trained Models & $\sigma=.12$ & $\sigma=.25$ &$\sigma=.25$ & $\sigma=.50$ \\
         \midrule
         Concave & $16\%$ & $6.71\%$ & $0\%$ & $0\%$\\
         Quasiconcave & $97\%$ & $98.17\%$ & $92.95\%$ & $90.47\%$\\
    \bottomrule
    \end{tabular}}
    \caption{We evaluate the generality of the concave and quasiconcave properties of the sigma-radius curve on two different datasets and four different trained models. We find that a significant number of data points do not exhibit a concave sigma-radius curve, while over $90\%$ of the data points demonstrate a quasiconcave sigma-radius curve.}
    \vspace{-0.3cm}
    \label{tab:generalityOfQC}
\end{table}

\subsection{Quasiconcavity}
This section lists the definition and two fundamental lemmas about quasiconcavity that will be used in designing our QCRS method.
Quasiconcavity is a generalization of concavity, defined as follows:
\begin{definition}
(quasiconcavity \cite{boyd2004convex}). 
A function $h: \mathbb{R}^n \rightarrow \mathbb{R}$ is quasiconcave if $\dom h$ is convex and for any $\theta \in [0,1]$ and $x, y$ $\in$ $\dom h$,
\begin{align*}
    h(\theta x +(1-\theta)y) \geq \min \{ h(x), h(y) \}.
\end{align*}
Furthermore, a function $h$ is strictly quasiconcave if $\dom h$ is convex and for any $x \neq y$, $x, y$ $\in$ $\dom h$, and $\theta \in (0,1)$:
\begin{align*}
    h(\theta x +(1-\theta)y) > \min \{ h(x), h(y) \}.
\end{align*}

\end{definition}
In this paper, we mainly use strict quasiconcavity.
Below, we list lemmas on strict quasiconcavity that we will use later.
\begin{lemma}\label{lma:global}
If a function $h$ is strictly quasiconcave, then any local optimal solution of $h$ must also be globally optimal.
\end{lemma}

\begin{lemma}\label{lma:left}
Suppose $h: \mathbb{R} \rightarrow \mathbb{R}$ is differentiable, and let $x^{*}$ be the optimal solution of $h$. The function $h$ is strictly quasiconcave if and only if the following two statements hold:
    $$\nabla h(x) > 0 , \forall x \in (-\infty,x^{*}) \text{ and } \nabla h(x) < 0 , \forall x \in (x^{*},\infty)$$
\end{lemma}
We defer the proofs of these two lemmas to Appendix D.

\subsection{Design}\label{sec:design}
In this section, we show quasiconcavity related to sigma-radius curves, i.e., $R(\sigma)$.
First, we differentiate $R(\sigma)$:
\begin{align*}
        \nabla_{\sigma} R(\sigma) = \frac{\partial R(\sigma)}{\partial \sigma} = \phiminus(\pabar) + \sigma \cdot \frac{\partial \phiminus(\pabar)}{\partial \sigma}
\end{align*}
Assume that $\sigma^*$ exists, where $\sigma^{*} = \text{arg max}_{\sigma} R(\sigma)$.
Then, according to Lemma~\ref{lma:left}, strict quasiconcavity is established if the gradient $\nabla_{\sigma} \phiminus$ adheres to the subsequent upper and lower bounds:
\begin{align*}
\begin{cases}
  \nabla_{\sigma} \phiminus =  \frac{\partial \phiminus(\pabar)}{\partial \sigma} > -\frac{\phiminus(\pabar)}{\sigma} & \text{for } \sigma < \sigma^* \\
  \nabla_{\sigma} \phiminus =  \frac{\partial \phiminus(\pabar)}{\partial \sigma} < -\frac{\phiminus(\pabar)}{\sigma} & \text{for } \sigma > \sigma^*
\end{cases}
\end{align*}
Intuitively, it indicates that $\nabla_{\sigma} \phiminus$ has a negative lower bound when $ \sigma < \sigma^*$ and a negative upper bound when $ \sigma > \sigma^*$.

The computation of a classifier's decision boundary is proven to be NP-hard, rendering the calculation of $\nabla_{\sigma} \phiminus$ impractical \cite{katz2017reluplex}.
Randomized smoothing employs a series of probabilistic approaches, such as MC sampling, to address this issue.
Similarly, instead of evaluating $\nabla_{\sigma} \phiminus$, we define a probabilistic condition based on Lemma~\ref{lma:left} to determine quasiconcavity of $R(\sigma)$: 
\begin{definition}\label{defn:qcassu}(($\upsilon^-,\upsilon^+$)-SQC condition)
Given an optimal $\sigma^*$, we call the sigma-radius curve satisfies ($\upsilon^-,\upsilon^+$)-Strict Quasiconcave Condition (($\upsilon^-,\upsilon^+$)-SQC condition), if for any $\{ \sigma | R(\sigma)>0 \}$ , $\nabla R(\sigma)$ satisfies the following: 
    \begin{align*}
        \mathop{Pr}\limits_{\sigma < \sigma^{*}}[\nabla R (\sigma)>0] = \upsilon^-, \quad \mathop{Pr}\limits_{\sigma > \sigma^{*}}[\nabla R (\sigma)<0] = \upsilon^+.
    \end{align*}
\end{definition}
Intuitively, if $\upsilon^-=\upsilon^+=1$, the condition states that the slope of the sigma-radius curve is positive on the left side of the optimal solution and negative on the right side.
Note that this condition is weaker and more general than the concentration assumption used in \cite{li2022double}, which requires additional assumptions on the distribution of data points.
It is also weaker than the concavity assumption used in DDRS \cite{alfarra2020data}.
Since the ($\upsilon^-,\upsilon^+$)-SQC condition is weak, we expect that more data points would satisfy this assumption.
Thus, we conduct some experiments on CIFAR-10 and ImageNet, which are detailed in Appendix B.
For each data point in CIFAR-10 and ImageNet, we first employ grid search to determine the optimal sigma $\sigma^*$. Subsequently, we uniformly sample 20 distinct sigma values and evaluate the ($\upsilon^-,\upsilon^+$)-SQC condition. 
In terms of concavity, we evaluate the Hessian of these 20 points.
The results are illustrated in Table~\ref{tab:generalityOfQC}. 
Taking the model with $\sigma=.25$ on CIFAR=10 as an example, we derive the following conclusions:
1) There are at most $98.17\%$ data points satisfy the ($1,1$)-SQC condition, while only a maximum of $6.7\%$ satisfy the concavity assumption;
2) Among these $98.17\%$ data points, their $\upsilon^-$ and $\upsilon^+$ are both within the interval $[0.8609,1]$ using $95\%$ confidence level, determined by the Clopper-Pearson interval. That is, these data points at least satisfy ($0.86,0.86$)-SQC condition. Notably, in our experiment, the mean values of $\upsilon^-$ and $\upsilon^+$ are greater than $0.99$;
3) If we set $\upsilon^-$ and $\upsilon^+$ to $0.99$, we can expect that $95\%$ of data points will achieve the optimal sigma, as the proposed method requires five iterations to converge ($0.99^5\approx0.95$). We will discuss the convergence in the next section.

\begin{figure*}[t]
\centering
\subfloat[$\sigma=0.12$]{\includegraphics[width=0.3\linewidth]{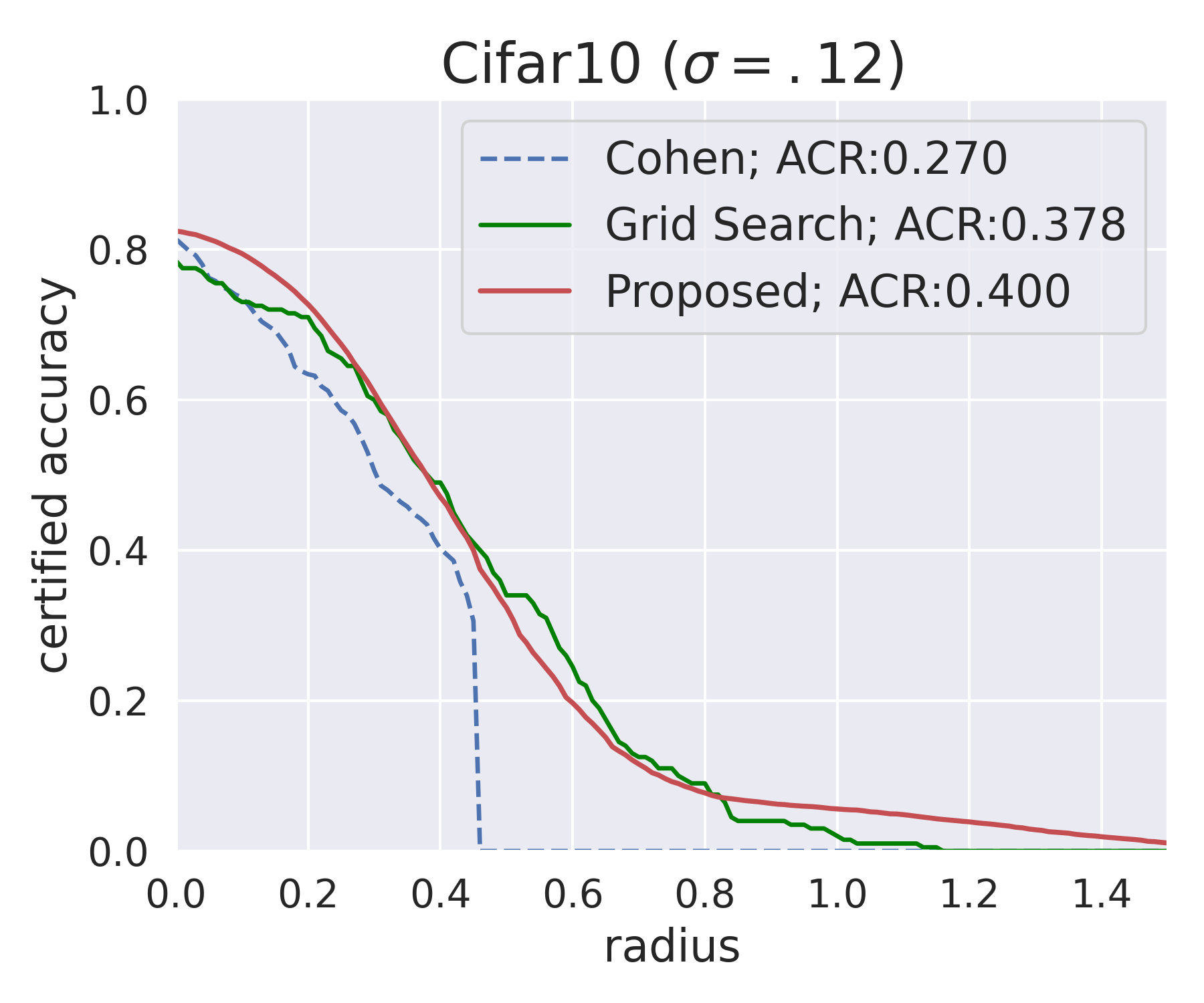}%
}
\subfloat[$\sigma=0.25$]{\includegraphics[width=0.3\linewidth]{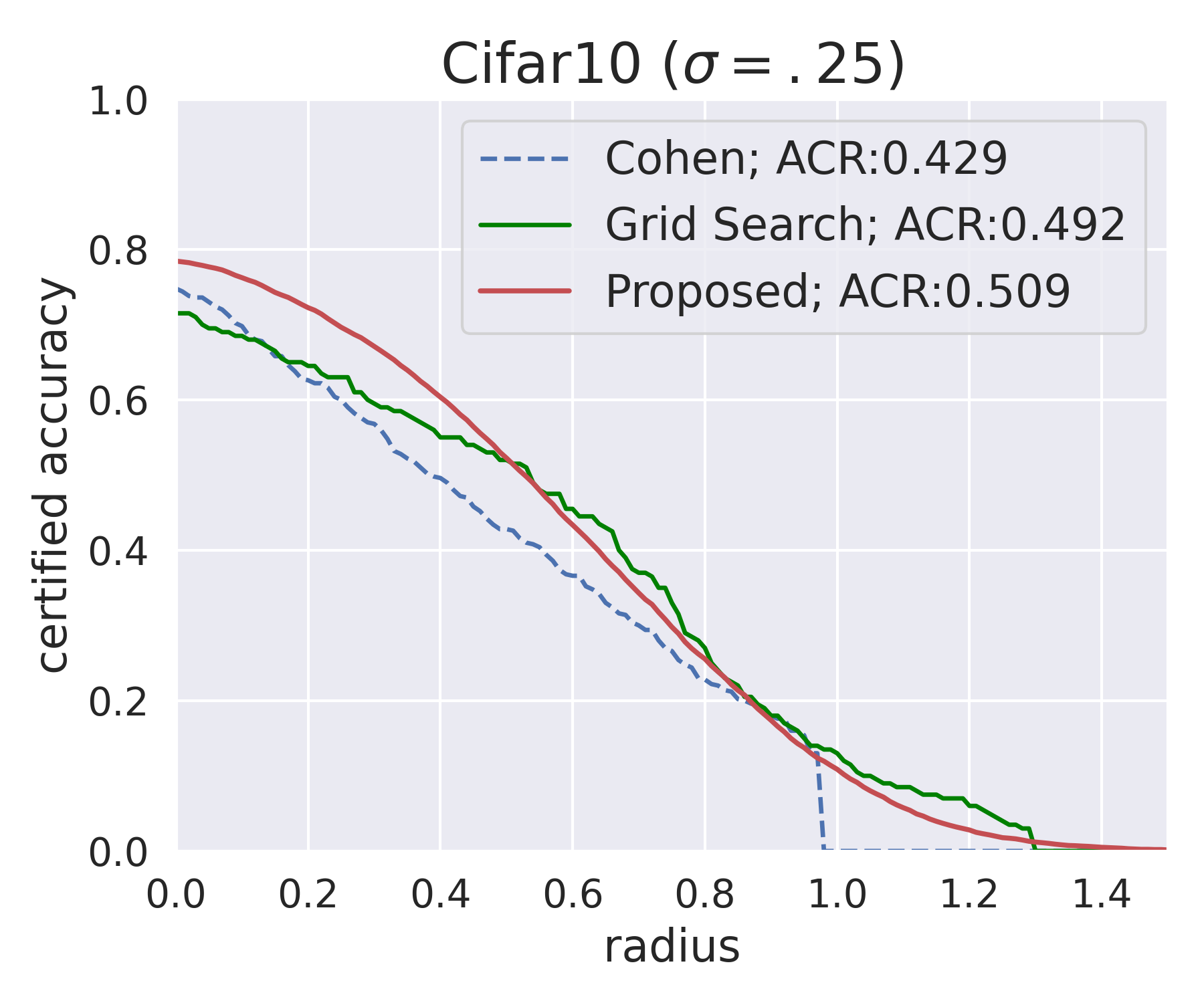}%
}
\subfloat[$\sigma=0.50$]{\includegraphics[width=0.3\linewidth]{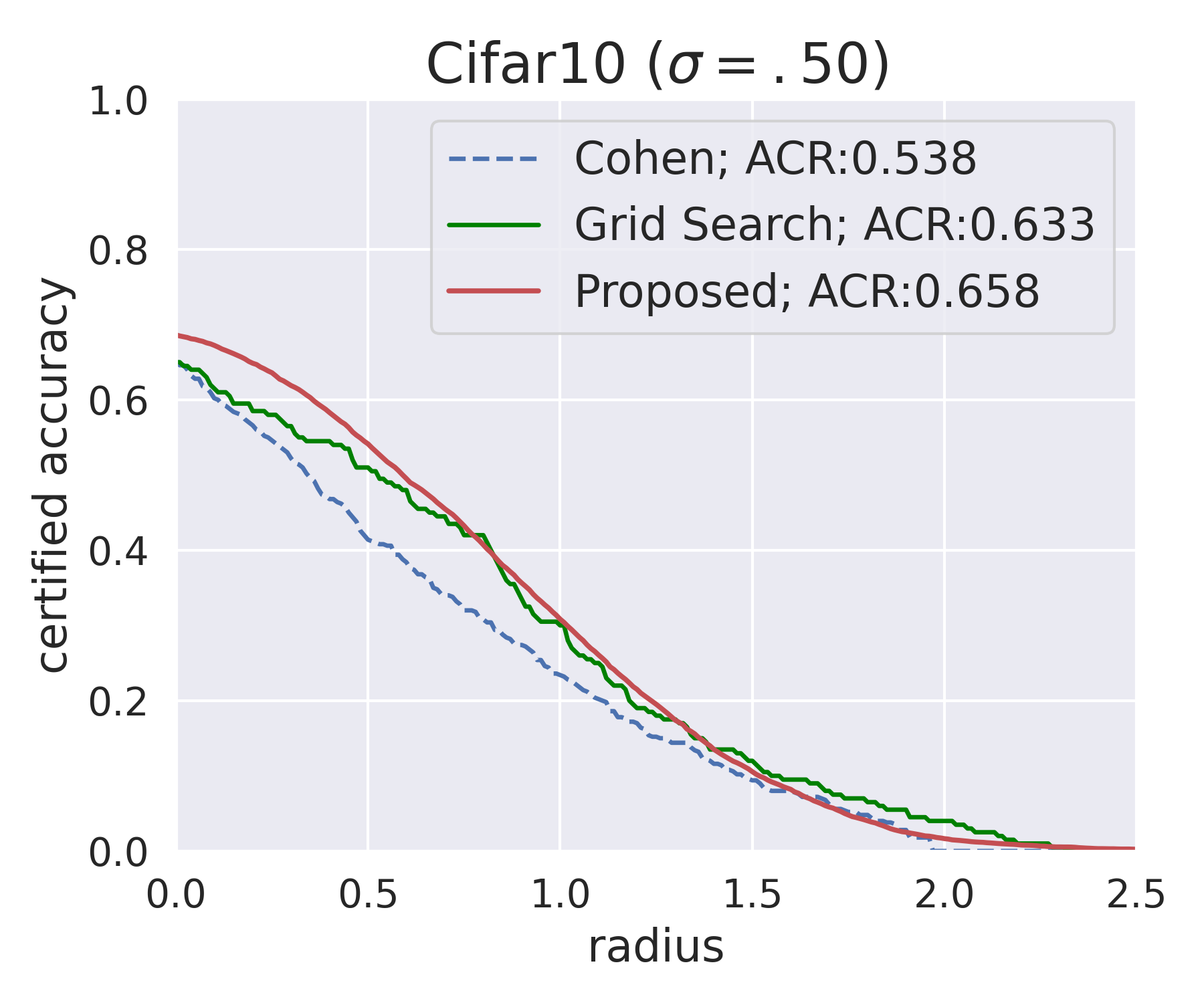}%
}
\vspace{-0.2cm}
\caption{The comparison between \textsc{Cohen}, grid search, and the proposed QCRS on the CIFAR-10 dataset. The models are trained by Gaussian augmentation with sigma (a) $0.12$, (b) $0.25$, and (c) $0.50$. The proposed QCRS outperforms the baseline method and is very close to grid search. In addition, we can observe the truncation effect on the curves of \textsc{Cohen}. }\label{fig:cifar_results}
\vspace{-0.3cm}
\end{figure*}

\newcommand{\mainalg}{
\begin{algorithm}[t]
\caption{Our proposed QCRS method}
\label{alg:main}
\textbf{Input}: Searching region $\sigma_{max}$ and $\sigma_{min}$; suboptimal interval $\varepsilon$; original sigma $\sigma_{0}$; gradient step $\tau$  \\
\textbf{Parameter}: momentum $M \leftarrow 0$ \\ 
\textbf{Output}: The optimal $\sigma$%
\begin{algorithmic}[1] 
\WHILE{$\sigma_{max} - \sigma_{min} > \varepsilon$}
    \STATE $\sigma \leftarrow ({\sigma_{min}+\sigma_{max}})/{2}$
    \STATE Calculate the gradient:
    \STATE $\qquad \nabla_{\sigma} R(\sigma) \leftarrow R(\sigma+\tau) - R(\sigma-\tau)$
    \IF {$sign(\nabla_{\sigma} R(\sigma))>0$}
    \STATE $\sigma_{min} \leftarrow \sigma$; $M \leftarrow 1$
    \ELSIF {$sign(\nabla_{\sigma} R(\sigma))<0$}
    \STATE $\sigma_{max} \leftarrow \sigma$; $M \leftarrow -1$
    \ELSE 
        \IF{$M \geq 0$}
            \STATE $\sigma_{max} \leftarrow \sigma$; $M \leftarrow -1$
        \ELSE 
            \STATE $\sigma_{min} \leftarrow \sigma$; $M \leftarrow 1$
        \ENDIF
    \ENDIF
\ENDWHILE
\STATE $\hat{\sigma} \leftarrow ({\sigma_{min}+\sigma_{max}})/{2}$
\RETURN $\sigma \leftarrow \argmax_{\sigma \in \{\hat{\sigma}, \sigma_{0}\}} R(\sigma)$
\end{algorithmic}
\end{algorithm}
}

We assume that a data point satisfies ($\upsilon^-,\upsilon^+$)-SQC condition, with the corresponding $\upsilon^-$ and $\upsilon^+$ being close to one.
According to Lemma~\ref{lma:left}, if we detect that the gradient of a point is positive, we can assert that the unique optimal sigma is on its right hand side.
Based on these rules, we design a time-efficient algorithm that can achieve optimal $\sigma$, shown in Algorithm~\ref{alg:main}.
Algorithm~\ref{alg:main} finds the optimal sigma efficiently based on binary search, and the sigma is the globally optimal solution, as demonstrated by Lemma~\ref{lma:global}.
On the other hand, the sigma values within the non-certified interval $\{ \sigma | R(\sigma)=0 \}$ must not be the solution. 
The gradients $\nabla R(\sigma)$ are likely to be zero in the interval because the curve is a horizontal line with $R(\sigma)=0$ there.
This leads to a gradient vanishing issue in Algorithm~\ref{alg:main}.
To circumvent this issue, we utilize momentum $M$ to guide the optimization direction.
Algorithm~\ref{alg:main} guarantees to find the same optimal solution as grid search if the curve satisfies ($1,1$)-SQC condition.
The time complexity is $N$ for grid search and $\log N$ for Algorithm~\ref{alg:main}, where $N$ is the number of points on the grid.
Therefore, the proposed method is significantly faster than grid search, while both of them can achieve the same optimal $\sigma$.

\mainalg

Prior work utilizes backpropagation to compute gradients, which is time-consuming, and the computed gradient is unstable due to MC sampling.
Therefore, we use forward passes to compute gradient, which takes the difference of two neighboring points.
This is because we only care about the gradient \textit{sign} rather than the exact value.
On the last stage of Algorithm~\ref{alg:main}, we employ a rejection policy that compares the resulting $\sigma$ to the original $\sigma$ and returns the larger one.

The proposed method is time-efficient compared to Insta-RS \cite{chen2021insta} and DDRS \cite{alfarra2020data}.
DDRS uses a low MC sampling number (one or eight) due to expansive computation and may obtain unstable gradients.
To verify this, we analyze the value of gradient under different MC sampling number.
The detailed results can be found in the Appendix H.
The results reveal that the gradient values vary dramatically when using low MC sampling numbers.
Thus, a low MC sampling number can not accurately estimate gradients, which would affect the gradient-based optimization.
However, the proposed QCRS only uses the gradient sign, which is more stable than the gradient value.
We observe that the sign of the gradient is consistent and hardly changes when the MC sampling number exceeds $500$.

\subsection{Convergence Analysis}
The proposed QCRS enjoys a convergence guarantee with the convergence rate analyzed below.
\begin{theorem}\label{thm:convergence_alg1}
Suppose (1,1)-SQC condition holds. Given hyper-parameters $\sigma_{min}$ and $\sigma_{max}$, let $\sigma_{t}$ be the $\sigma$ value after $t$ iterations in Algorithm~\ref{alg:main}. Algorithm~\ref{alg:main} converges to optimal $\sigma^*$ as follows:
\begin{align*}
    \frac{ \sigma_{max} - \sigma_{min}}{2^t} \geq | \sigma_t - \sigma^*|.
\end{align*}
\end{theorem}
Theorem~\ref{thm:convergence_alg1} means the convergence rate of Algorithm~\ref{alg:main} is $\mathcal{O}((\frac{1}{2})^t)$. 
We defer the proof to Appendix D.

On the other hand, gradient-descent-based method requires much stricter assumptions, such as L-smoothness and $\mu$-strongly concavity, to guarantee convergence.
When these conditions are satisfied, the convergence rate is as below \cite{nesterov2018lectures}:
\begin{align*}
|\sigma_t - \sigma^{*}|^2 \leq (\frac{L-\mu}{L+\mu})^{(t-1)
}|\sigma_1 - \sigma^{*}|^2.
\end{align*}
The convergence rate is $\mathcal{O}((\frac{L-\mu}{L+\mu})^t)$, where $L$ and $\mu$ depend on the data points.
Note that only a maximum of $6.7\%$ of data points satisfy the concavity assumption empirically, and even fewer data points satisfy L-smoothness and $\mu$-strongly concavity simultaneously.
Thus, some concave optimization methods cannot converge to the optimal sigma values for most data points.


\begin{table*}[t]
\begin{center}
\scalebox{0.89}{
\begin{tabular}{l|cccc}
\toprule
\multicolumn{1}{l|}{ACR}  &\multicolumn{1}{c}{$\sigma=.12$} &\multicolumn{1}{c}{$\sigma=.25$} &\multicolumn{1}{c}{$\sigma=.50$} &\multicolumn{1}{c}{Time Cost (Sec.)}
\\ 
\hline
\cohen         & 0.270 & 0.429 & 0.538 & 6.50$\pm$0.021 \\
DDRS \cite{alfarra2020data} & 0.310 & 0.448 & 0.540 & 7.39$\pm$0.016 \\
Grid Search   & 0.378 & 0.492 & 0.633 & 155.80$\pm$0.50\\
\textbf{QCRS}      & \textbf{0.400} & \textbf{0.509} & \textbf{0.658} & 6.96$\pm$0.017\\
\bottomrule
\end{tabular}}
\vspace{-0.2cm}
\caption{ACR and Time Cost for CIFAR-10.}
\label{table:cifar10}
\vspace{-0.3cm}
\end{center}
\end{table*}

\section{Experimental Results}
We evaluate the proposed QCRS and present the experimental results on CIFAR-10 \cite{krizhevsky2009learning} and ImageNet \cite{russakovsky2015imagenet}.
We also verify that QCRS can be combined with
training-based techniques like MACER \cite{zhai2019macer} to produce state-of-the-art certification results.
Detailed implementation information is available in Appendix E.\footnote{Code: \url{https://github.com/ntuaislab/QCRS}}
We also present additional experiments in Appendix, including an ablation study, error analysis, and more.
Following \cite{zhai2019macer}, we use the average certified radius (ACR) as the performance metric, defined as:
$
    ACR = \frac{1}{|\mathcal{D}_{test}|}\sum_{x \in \mathcal{D}_{test}} R(x,y; g),
$
where $\mathcal{D}_{test}$ is the test dataset, and $R(x,y; g)$ is the certified radius obtained by the smoothed classifier $g$.

\subsection{CIFAR-10 and ImageNet}

\begin{table*}[!t]
    \centering
    \scalebox{0.86}{
    \begin{tabular}{l|ccccccccc}
    \toprule
         & \multicolumn{9}{c}{Certified Accuracy}\\
        Certified radii $R$ & $0.25$& $0.5$& $0.75$& $1.0$& $1.25$& $1.5$& $1.75$& $2.0$& $2.25$\\
        \midrule
        \cohen & 0.55 & 0.41 & 0.32 & 0.23 & 0.15 & 0.09 & 0.05 & 0.00 & 0.00\\
        DDRS \cite{alfarra2020data} & 0.56 & 0.44 & 0.34 & 0.23 & 0.15 & 0.08 & 0.04 & 0.00 & 0.00\\
        DSRS \cite{li2022double} & 0.55 & 0.42 & 0.30 & 0.19 & 0.09 & 0.04 & 0.00 & 0.00 & 0.00\\
        Grid Search (24 points) & 0.58 & 0.51 & 0.42 & 0.30 & 0.18 & \textbf{0.12} & \textbf{0.07} & \textbf{0.04} & 0.01\\
        \textbf{QCRS (Proposed)} & \textbf{0.64} & \textbf{0.54} & \textbf{0.43} & \textbf{0.31} & \textbf{0.20} & 0.11 & 0.05 & 0.02 & 0.01\\
    \bottomrule
    \end{tabular}}
    \vspace{-0.2cm}
    \caption{Certified accuracy under different radii $R$ of DSRS, DDRS, Grid Search, and the proposed QCRS.}
    \label{tab:dsrs}
    \vspace{-0.4cm}
\end{table*}

Fig.~\ref{fig:cifar_results} compares the radius-accuracy curves for different methods on the CIFAR-10 dataset.
In the figure, we also show the corresponding ACR, which is the area under the radius-accuracy curve.
Table~\ref{table:cifar10} presents the ACR of different methods and their runtime cost.
The proposed QCRS outperforms the original randomized smoothing method \cite{cohen2019certified} by significant margins: $48\%$, $18\%$, and $22\%$ for the models trained with $\sigma = \{0.12, 0.25, 0.50\}$, respectively. 
The main performance gain comes from reducing the truncation effect (the waterfall effect) on the radius-accuracy curve.
We also compare QCRS to grid search.
Since grid search is extremely computationally expensive, we only test the images with $id = 0,50,100,...,9950$ in CIFAR-10.
Despite using 24 points in the grid search, which costs approximately $24$ times more in runtime than QCRS, QCRS still outperforms grid search.
This is due to QCRS being more time-efficient, which enables a broader and more accurate search region compared to grid search.
Furthermore, QCRS guarantees to achieve the same optimal as grid search if the ($1,1$)-SQC condition holds.
Regarding the computational cost, the proposed method only takes about $7\%$ additional inference time compared to the original \cohen method, as shown in Table~\ref{table:cifar10}.

We also compare the proposed QCRS with two state-of-the-art randomized smoothing methods, DSRS \cite{li2022double} and DDRS \cite{alfarra2020data}. 
We follow their setting to evaluate the proposed method on CIFAR-10 for fair comparisons, and defer other minor comparisons and analyses to Appendix C.
As Table~\ref{tab:dsrs} shows, for the certified accuracy under radius at $0.5$, DSRS and DDRS improve \cohen by $2.4\%$ and $7.3\%$, respectively.
On the other hand, the proposed QCRS improves \cohen by $31.7\%$.
Therefore, among the methods that boost certified radii, QCRS improves \cohen most effectively.

\begin{figure}[t]
\centering
\subfloat[$\sigma=0.25$]{\includegraphics[width=0.46\linewidth]{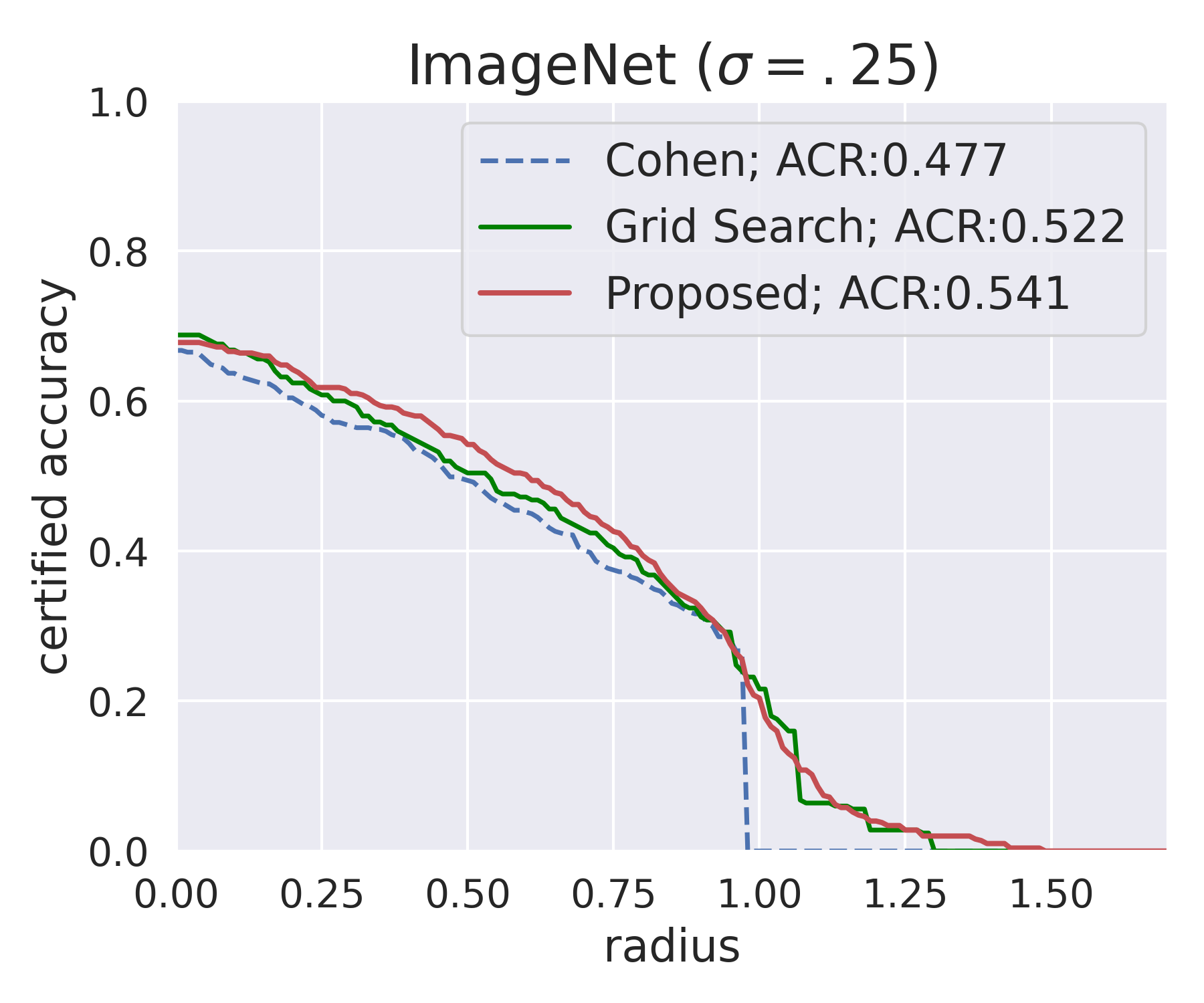}%
}
\subfloat[$\sigma=0.50$]{\includegraphics[width=0.46\linewidth]{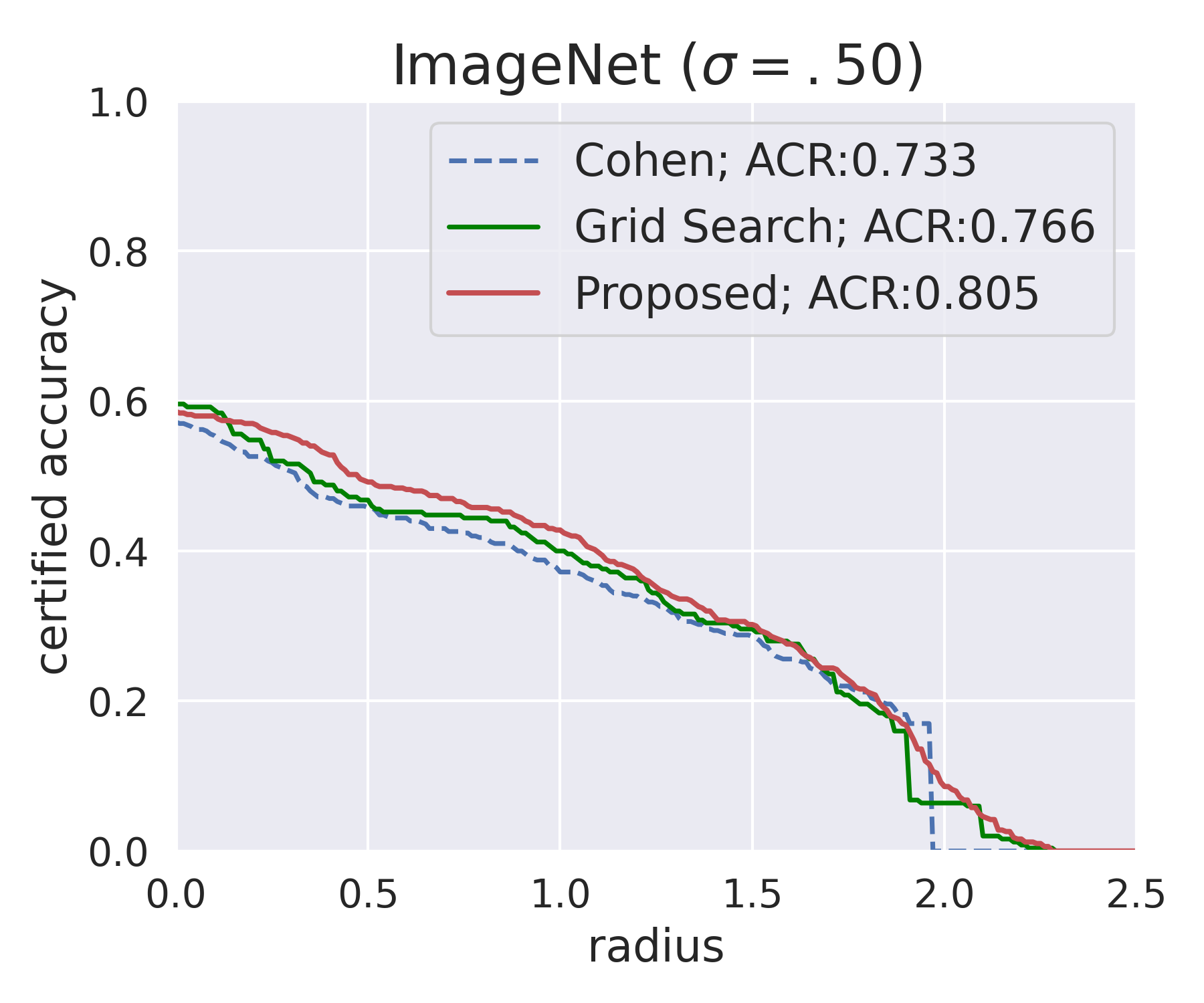}%
}
\vspace{-0.2cm}
\caption{The comparison between \textsc{Cohen}, grid search, and QCRS on ImageNet. Following Cohen, we only use $500$ images in the validation set. The models are trained by Gaussian augmentation with $\sigma=$ (a) $0.25$ and (b) $0.50$.}
\label{fig:imagenet_results}
\vspace{-0.3cm}
\end{figure}

Fig.~\ref{fig:imagenet_results} shows the results on ImageNet. 
Following \textsc{Cohen}, only $500$ images with $id=0,100,200,...,49900$ in the validation set were used.
We avoided using the $\sigma=1.0$ model as \cite{mohapatra2021hidden} revealed that a large $\sigma$ leads to serious fairness concerns, necessitating a restriction on $\sigma$ in randomized smoothing. 
Additionally, we observed the similar fairness issue in \cite{mohapatra2021hidden}, to be discussed later.
For the model with $\sigma=.25$, the proposed method improves ACR from $0.477$ to $0.541$, roughly $13.4\%$.
Similarly, for the model with $\sigma=.50$, the proposed method improves ACR from $0.733$ to $0.805$, roughly $9.8\%$.
In addition, the proposed method overcomes the truncation effect, providing a larger certified radius than \textsc{Cohen}.
As for the grid search, similar to CIFAR-10, it is computationally expensive, so we set the number of searching points to be $11$ on ImageNet.
As mentioned earlier, although the grid search can provide the optimal certified radius if the cost does not matter, its searching region and precision are limited in practical application.
That is why the proposed method achieves a slightly superior ACR than the brute-force grid search method in Fig.~\ref{fig:imagenet_results}, while the runtime is roughly 11 times faster than it.

\subsection{MACER}

\begin{figure}[t]
\centering
\subfloat[$\sigma=0.25$]{\includegraphics[width=0.46\linewidth]{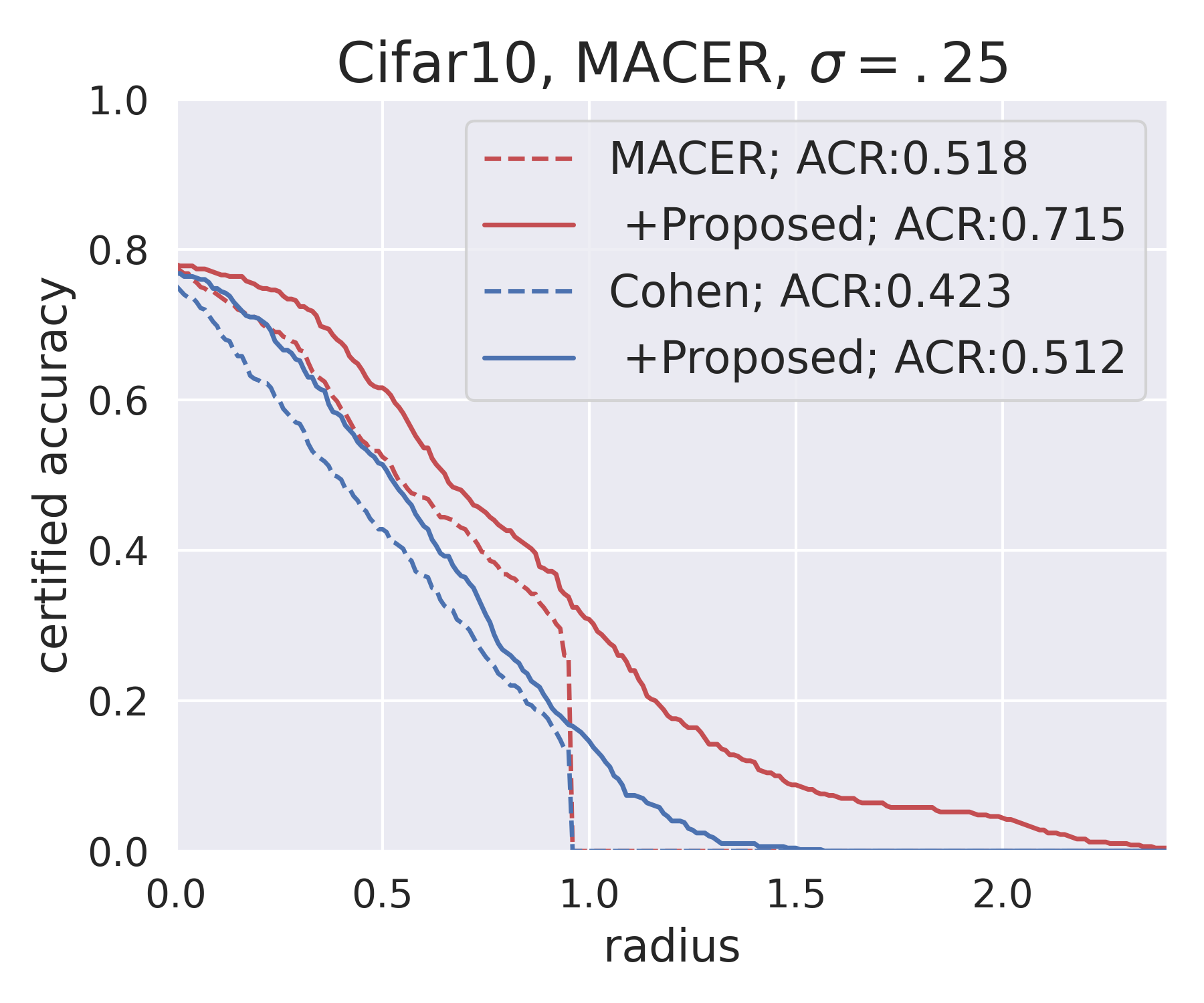}%
}
\subfloat[$\sigma=0.50$]{\includegraphics[width=0.46\linewidth]{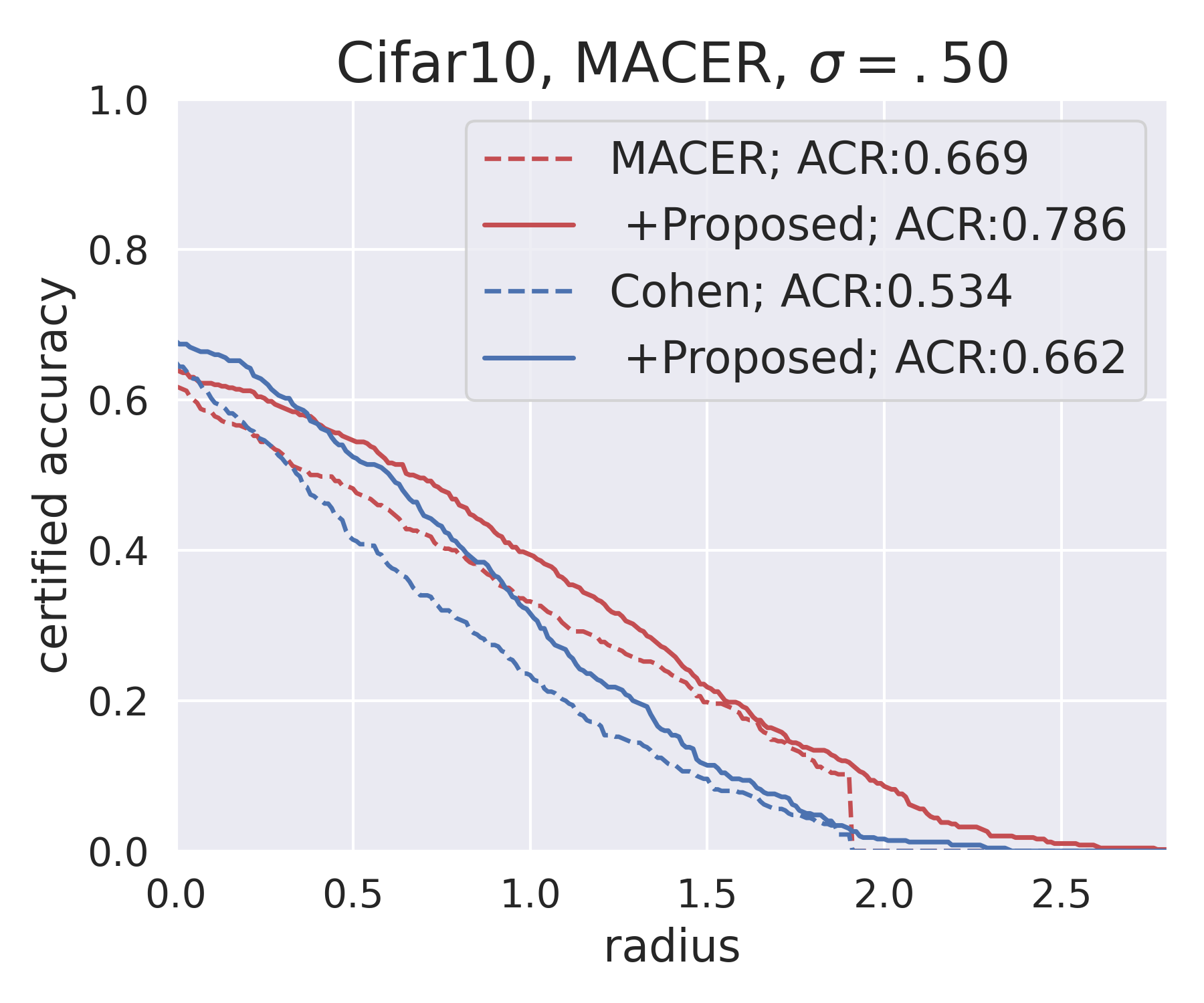}%
}
\vspace{-0.2cm}
\caption{The performance of QCRS incorporating training-based methods. We use MACER model with (a) $\sigma=0.25$ and (b) $\sigma=0.50$. Both QCRS and MACER provide similar improvements over \textsc{Cohen}, but QCRS incurs little computational overhead. Combining QCRS and MACER provides state-of-the-art certified radii.}
\label{fig:macer_results}
\vspace{-0.4cm}
\end{figure}

The proposed method focuses on enhancing randomized smoothing while building the smoothed classifier. 
Thus, it is orthogonal to the approach that aims to boost certified radii during training stage.
QCRS can incorporate with training-based methods.
The most representative training-based method to enhance certified radius is MACER.
We apply the proposed method to the models trained by MACER and observe significant improvement in terms of the certified radius.
Fig.~\ref{fig:macer_results} illustrates the results of radius-accuracy curves, and Table~\ref{tab:macer} shows the detailed comparison.
As Table~\ref{tab:macer} shows, for the model trained by $\sigma=.25$, \cohen achieves $0.423$ ACR, and MACER enhances this ACR to $0.518$, roughly $22.5\%$.
Next, our proposed QCRS improves MACER ACR from $0.518$ to $0.715$, roughly $38\%$.
Thus,  QCRS and MACER together can significantly boost the original Cohen's RS roughly $69\%$.
Similarly, for the model trained by $\sigma=.50$, QCRS and MACER enhance Cohen's RS from $0.534$ to $0.786$, approximately $+47.2\%$.
Furthermore, Table~\ref{tab:macer} also includes the results of DDRS incorporating MACER. 
It can be observed that MACER has a positive impact on the performance of the inference-phase randomized smoothing methods.
Among the methods evaluated, QCRS incorporating MACER achieves the state-of-the-art performance.
On the other hand, QCRS and MACER improves \cohen to $0.512$ and $0.518$, respectively.
That is, QCRS can enlarge the certified radius to the extent that MACER does, but it does not need any training procedure.
As datasets become larger and larger today, re-training may be computationally prohibited. 
Thus, QCRS benefits from its efficient workflow that enlarges the certified radius with negligible cost.

\newcommand{\macertableI}{%
    \scalebox{0.98}{
    \begin{tabular}{l|cc}
    \toprule
        ~ &  \multicolumn{2}{c}{Training}   \\
        Test & \cohen & MACER  \\ \hline
        \cohen & 0.423 & 0.518  \\
        DDRS  & 0.448 ($+6\%$) & 0.561 ($+8\%$) \\
        \textbf{QCRS} & 0.512 ($+21\%$) & \textbf{0.715} ($+38\%$) \\
    \bottomrule
    \end{tabular}}
}

\newcommand{\macertableII}{%
    \scalebox{0.98}{
    \begin{tabular}{l|cc}
    \toprule
        ~ &  \multicolumn{2}{c}{Training}   \\ %
        Test & \cohen & MACER  \\ \hline
        \cohen & 0.534 & 0.669  \\
        DDRS  & 0.540 ($+1\%$) & 0.702 ($+5\%$)\\
        \textbf{QCRS} & 0.662 ($+24\%$) & \textbf{0.786} ($+18\%$)  \\ 
    \bottomrule
    \end{tabular}}
}

\begin{table}[t]%
  \centering
  \subfloat[][$\sigma=0.25$]{\macertableI}%
  \quad
  \subfloat[][$\sigma=0.50$]{\macertableII}
  \vspace{-0.2cm}
  \caption{The ACR results of different inference-phase RS incorporating different training-phase RS. The training-phase RS are \cohen or MACER with (a) $\sigma=.25$ and (b) $\sigma=.50$. The inference-phase RS are \textsc{Cohen}, DDRS, or QCRS. The table also includes the improvement ratios compared to \textsc{Cohen}. }%
  \label{tab:macer}%
  \vspace{-0.3cm}
\end{table}

\subsection{Computational Cost}
We briefly analyze the computational cost compared to \textsc{Cohen}.
The sigma searching region of Algorithm~\ref{alg:main} is $0.5-0.12=0.38$.
Because the convergence rate of Algorithm~\ref{alg:main} is $\frac{ \sigma_{max} - \sigma_{min}}{2^t} \geq | \sigma_t - \sigma^*|$, if $t \geq 6$, we can achieve $0.006$-optimal, i.e., $|\sigma - \sigma^*| < 0.006$.
As for the gradient computation, it requires $1,000$ forward propagations for each iteration.
Thus, we roughly require additional $6,000$ forward propagations for each data point to achieve the optimal sigma value.
The standard RS needs $100,000$ forward propagations, so the overhead of the proposed QCRS is approximately $6\%$.
Empirically, we observe an approximately $7\%$ overhead as Table~\ref{table:cifar10} illustrates.

We compare the computational cost of QCRS with those of DDRS \cite{alfarra2020data}, DSRS \cite{li2022double}, and Insta-RS \cite{chen2021insta}.
The DDRS method adopts the radius as its objective function and utilizes gradient descent to directly optimize the radius. This approach involves computing the gradient of the radius multiple times to update the sigma value, which requires performing multiple back propagations. As shown in Table~\ref{table:cifar10}, certifying an image using DDRS takes roughly $7.39$ seconds. This incurs an overhead of $14\%$, which is twice as high as the proposed QCRS method.
As for DSRS, in our experiments, it incurs roughly $100\%$ overhead compared to \textsc{Cohen}.
Finally, Insta-RS employs multi-start gradient descent, so its computational cost is the highest among the methods considered in this paper.

\subsection{Constant Sigma during Deployment}
A consistent sigma may be required during deployment, presenting a limitation for input-specific RS. 
The memory bank strategy proposed by DDRS \cite{alfarra2020data}, compatible with our QCRS, can address this issue. 
It is crucial to understand QCRS still can apply a consistent sigma during deployment, and it does not undermine the soundness of certification, though it may limit its practical utility in real-world scenarios. 
Specifically, in binary scenarios, the proposed method enables the customization of an optimal sigma for specific classes, thereby effectively enhancing their certified radii.
For further details, please see Appendix J.

\subsection{Fairness Issue}
As discussed in \cite{mohapatra2021hidden}, randomized smoothing suffers from fairness issue.
The unbounded class in the data space dominates as $\sigma$ increases.
We investigate the optimal $\sigma$ for each class in CIFAR-10.
As Table~\ref{tab:fairness} illustrates, class ``cars'' exhibits a larger optimal sigma value compared to the other classes. 
This discrepancy arises from the possibility that ``cars'' represents the unbounded class; thus, an increase in the sigma value leads to a continuous expansion of the radius.
Exploring potential solutions to address this issue remains an intriguing direction.
\begin{table}[tp]
    \centering
    \scalebox{0.75}{
    \begin{tabular}{ccccc}
         \toprule
         airplanes & cars & birds & cats & deer \\
         $0.53 \pm 0.21$ & $\mathbf{0.84 \pm 0.41}$ & $0.37 \pm 0.10$ & $0.36 \pm 0.11$ & $\mathbf{0.36 \pm 0.10}$\\
         \midrule
         dogs & frogs & horses & ships & trucks \\
         $0.50 \pm 0.17$ & $0.53 \pm 0.08$ & $0.53 \pm 0.12$ & $0.42 \pm 0.08$ & $0.61 \pm 0.14$\\
         \bottomrule
    \end{tabular}}
    \vspace{-0.2cm}
    \caption{The optimal sigma values for different classes in CIFAR-10. The base model is trained with $\sigma = 0.5$. }
    \label{tab:fairness}
    \vspace{-0.3cm}
\end{table}

\section{Conclusion}
In this work, we exploit and empirically demonstrate the quasiconcavity of the sigma-radius curve.
The ($\upsilon^-,\upsilon^+$)-SQC condition is general and easy to satisfy.
Therefore, most data points (approximately $99\%$) conform to this condition.
Based on the ($\upsilon^-,\upsilon^+$)-SQC condition, we develop an efficient input-specific method called \textbf{QCRS} to efficiently find the optimal $\sigma$ used for building the smoothed classifier, enhancing the traditional randomized smoothing significantly.
Unlike former inference-time randomized smoothing methods that suffer from marginal improvement or high computational overhead, the proposed method enjoys better certification results and lower cost.
We conducted extensive experiments on CIFAR-10 and ImageNet, and the results show that the proposed method significantly boosts the average certified radius with $7\%$ overhead.
QCRS improves ACR, overcoming the trade-off on the radius-accuracy curve and eliminating the truncation effect.
In addition, we combine the proposed QCRS with a training-based technique, and the results demonstrate the state-of-the-art average certified radii on CIFAR-10 and ImageNet.
A direction for future work is to generalize the proposed method to $\ell_p$ ball and different distributions. A better training approach for QCRS is also an interesting future research direction.

\section*{Acknowledgement}
This work was supported in part by the National Science
and Technology Council under Grants MOST 110-2634-F002-051, MOST 110-2222-E-002-014-MY3,  NSTC 113-2923-E-002-010-MY2, NSTC-112-2634-F-002-002-MBK, by National Taiwan University  under Grant NTU-CC-112L891006, and by Center of Data Intelligence: Technologies, Applications, and Systems under Grant NTU-112L900903.

\bibliography{ms}

\clearpage

\appendix 
\noindent{\Large \textbf{Appendix}}

\section{Quasiconcavity of Sigma-Radius Curves}
\begin{figure}[tp]
    \centering
    \includegraphics[width=0.85\linewidth]{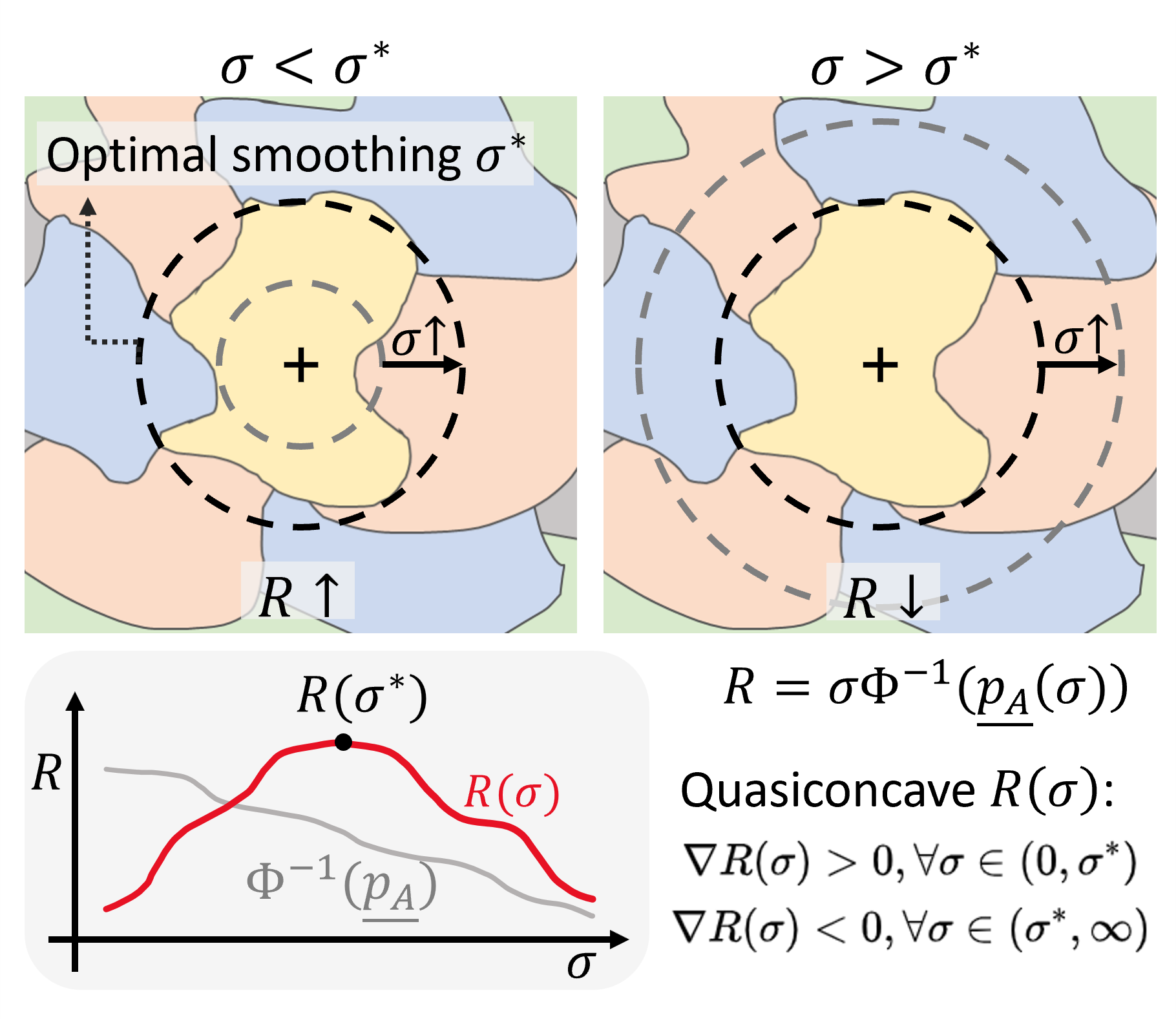}
    \caption{Illustration of the relation between $R$ and $\sigma$. 
As $\sigma$ gradually increases from zero, $\phiminus(\underline{p_A})$ does not change much as most sampled points are still within the boundary, leading to an increase in $R$. However, as $\sigma$ further increases, the other class catches up, and $\phiminus(\underline{p_A})$ decreases rapidly, causing $R$ to decrease. Thus, a negative correlation exists between $\phiminus(\underline{p_A})$ and $\sigma$. The objective is to identify the optimal value of $\sigma$ that maximizes the radius $R$.}
    \label{fig:concepts}
\end{figure}

In this section, our aim is to provide an intuitive explanation of the quasiconcavity observed on the sigma-radius curves.
First, we scrutinize the sigma-radius curve, denoted as $R(\sigma) = \sigma \cdot \phiminus(\pabar)$, from the perspective of the data space. 
Here, $\pabar$ signifies the lower bound of the Gaussian measure for class $A$ with a specific $\sigma$, and $\phiminus$ corresponds to the inverse Gaussian cumulative distribution function.
As depicted in Fig.~\ref{fig:concepts}, while $\sigma$ gradually increases from zero, $\pabar$ remains close to 1 because most of the measure falls within the boundary. Consequently, $R$ increases with the increment of $\sigma$. 
As $\sigma$ continues to increase, $\pabar$ begins to decrease, especially when the Gaussian kernel incorporates more samples from the wrong class. 
This results in a significant decline in $\phiminus(\pabar)$ and consequently causes $R$ to decrease.
Thus, there exists a negative relationship between $\sigma$ and $\phiminus(\underline{p_A})$, causing the radius initially to increases and then decreases as the sigma value varies from zero to infinity.
The objective is to identify the optimal value of $\sigma$ that maximizes the radius $R$.
This observation inspires us to leverage the quasiconcavity property of the problem structure to effectively find the optimal $\sigma$.

\section{Evaluating Quasiconcavity}
We conduct empirical evaluations to assess the concavity and the quasiconcavity of data points on CIFAR-10 and ImageNet datasets.

\noindent \textbf{Concavity:} To mitigate the computational burden, we sample 20 values of $\sigma$ and subsequently evaluate the objective function $R(\sigma)$ using 100,000 Monte Carlo sampling. The specific 20 sigma values will be presented later in this paper. Subsequently, we compute the \textit{Hessian} value for these 20 sampled points. It is crucial to note that for concave functions, the \textit{Hessian} of $R(\sigma)$ should be non-positive. Thus, any data point demonstrating a positive \textit{Hessian} is indicative of a non-concave function.

\noindent \textbf{Quasiconcavity:} The assessment of quasiconcavity leverages Lemma 2 as detailed in the main paper. Initially, we employ a grid search methodology to identify the optimal value for $\sigma$. Analogous to the concavity evaluation, we sample the same set of 20 sigma values and subsequently compute the gradient sign of $R(\sigma)$ using the forward method introduced in the main paper along with 100,000 Monte Carlo sampling. The fulfillment of the conditions stipulated in Lemma 2 is integral. Failure to satisfy Lemma 2 indicates that $R(\sigma)$ is not a quasiconcave function.

Note that in our evaluation, we provide a guarantee against false negatives, ensuring \textit{completeness}. Specifically, if a data point is identified as non-concave or non-quasiconcave, it is indeed non-concave or non-quasiconcave. However, it is important to acknowledge the potential existence of false positive samples. 
Therefore, Table~1 in the main paper provides an illustration of the upper bound for the number of samples that satisfy the criteria of concavity/quasiconcavity.
We demonstrate that concavity is challenging for real-world data, whereas quasiconcavity is potentially more prevalent.


To further investigate the generality of quasiconcavity, we use MC sampling to estimate $\upsilon^-$ and $\upsilon^+$. The sampling number is set to 20 for each data point, i.e., 20 different sigma values, and for each sigma value, we use 100,000 MC sampling to compute the radius.
On average, the estimated $\upsilon^-$ and $\upsilon^+$ for all data points, including both quasiconcave and non-quasiconcave cases, are $99.77$ and $99.11$ respectively, for CIFAR-10 and ImageNet datasets.
That is, approximately $0.9977^5 = 98.85\%$ of data points on CIFAR-10 are expected to converge to the optimal sigma, and for ImageNet, this percentage is $95.62\%$.

The experiment settings for Table~1 in the main paper are as follows: 
1) For the model that is trained on CIFAR-10 and $\sigma=0.12$, we sample the sigma value at 0.020, 0.030, 0.040, 0.050, 0.060, 0.070, 0.080, 0.090, 0.100, 0.110, 0.120, 0.130, 0.140, 0.150, 0.160, 0.170, 0.180, 0.190, 0.200, and 0.220;
2) For the model that is trained on CIFAR-10 and $\sigma=0.25$, we sample the sigma value at 0.15, 0.18, 0.2, 0.21, 0.22, 0.23, 0.24, 0.25, 0.26, 0.27, 0.28, 0.29, 0.3, 0.31, 0.32, 0.33, 0.35, 0.4, 0.45, and 0.5;
3) For the model that is trained on ImageNet and $\sigma=0.25$, we sample the sigma value at 0.100, 0.115, 0.130, 0.145, 0.160, 0.175, 0.190, 0.205, 0.220, 0.235, 0.250, 0.265, 0.280, 0.295, 0.310, 0.325, 0.340, 0.355, 0.370, 0.385, and 0.400;
4) For the model that is trained on ImageNet and $\sigma=0.25$, we sample the sigma value at 0.3, 0.32, 0.34, 0.36, 0.38, 0.40, 0.42, 0.44, 0.46, 0.48, 0.50, 0.52, 0.54, 0.56, 0.58, 0.60, 0.6,2 0.64, and 0.66. 
Note that we increase the sampling density around the $\sigma$ with which the model is trained.

\begin{figure}[t]
    \centering
    \includegraphics[width=0.45\textwidth]{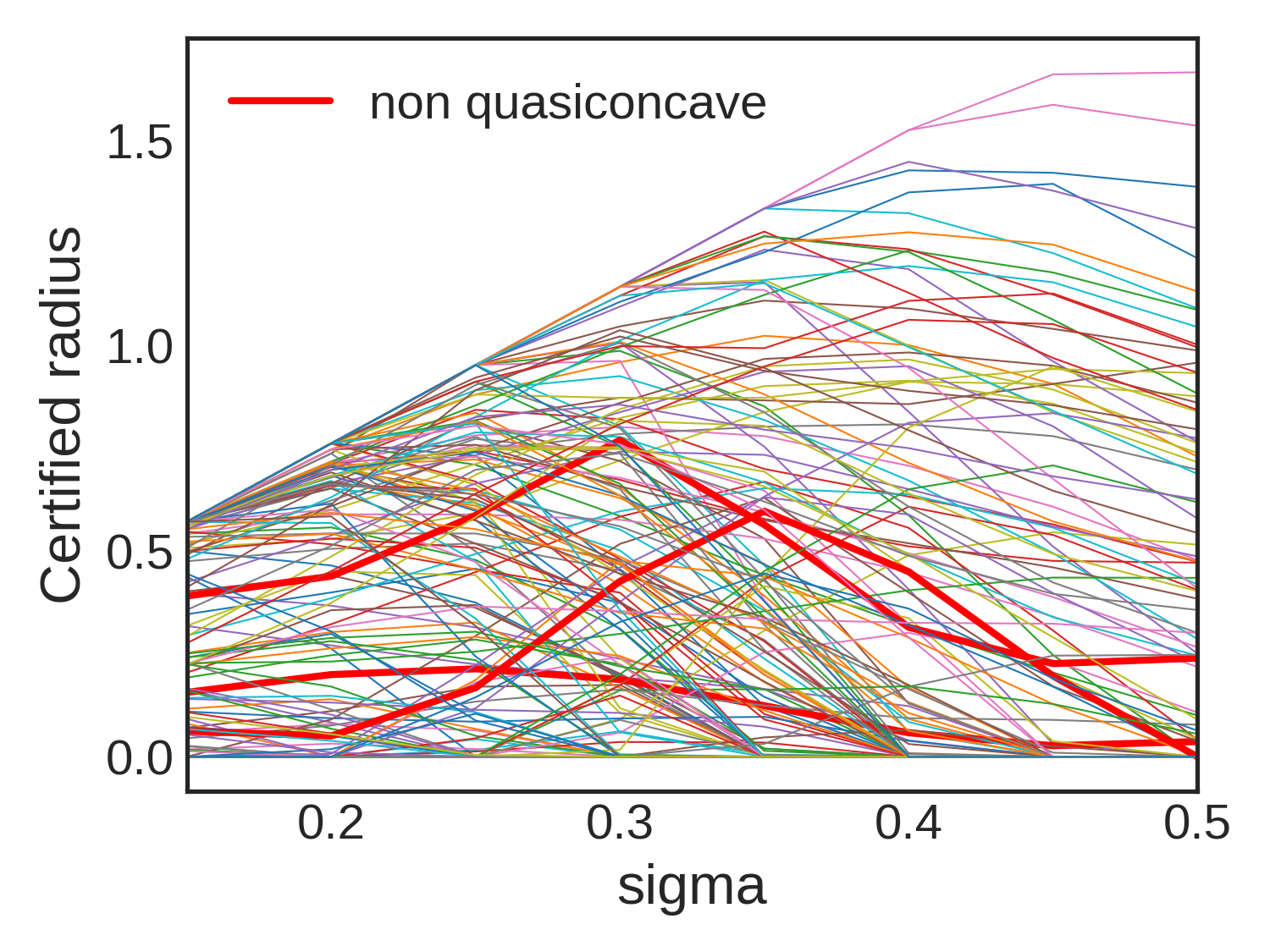}
    \caption{The sigma-radius curves on CIFAR-10. We can observe the quasiconcavity of the curves in this figure. The different curves indicate different data points. We evaluate the sigma-radius curves of $164$ images from CIFAR-10 that can be certified by \textsc{Cohen} and check their concavity and quasiconcavity numerically. $6.7\%$ and $98.2\%$ of data points are concave and quasiconcave, respectively. The bold red curve is non-quasiconcave, and the others are all quasiconcave.}
    \label{fig:sigma_curve}
\end{figure}

\begin{figure}[t]
    \centering
    \includegraphics[width=0.45\textwidth]{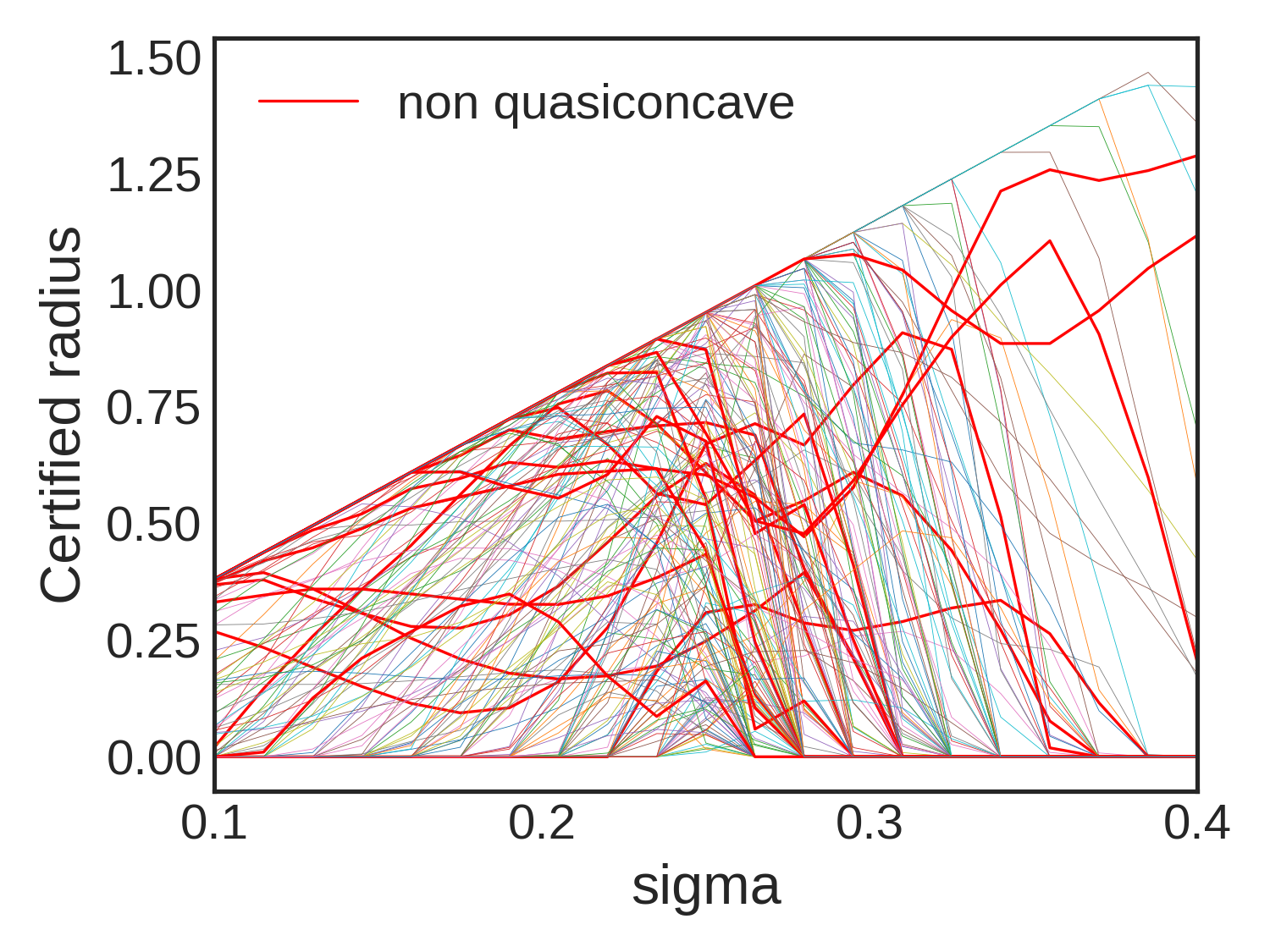}
    \caption{The sigma-radius curves on ImageNet. We can observe the quasiconcavity of the curves in this figure. The different curves indicate different data points. We evaluate the sigma-radius curves of $210$ images from ImageNet that can be certified by \textsc{Cohen} and check their concavity and quasiconcavity numerically. $0\%$ and $93\%$ of data points are concave and quasiconcave, respectively. The bold red curve is non-quasiconcave, and the others are all quasiconcave. Different from CIFAR-10, the count of non-quasiconcave and non-concave data points rises in the case of ImageNet. One plausible explanation is the heightened complexity of the data manifold for classifiers on ImageNet due to its higher number of classes and dimension. This complexity contributes to intricate and non-smooth decision boundaries.}
    \label{fig:sigma_curve_image}
\end{figure}

To visualize the sigma-radius curve, we plot the sigma-radius curves in Fig.~\ref{fig:sigma_curve}, which was derived from the CIFAR-10 model with $\sigma=0.25$. The figure clearly illustrates that only three curves fail to satisfy the quasiconcave condition, represented by the bold red curve.
Furthermore, Fig.~\ref{fig:sigma_curve_image} shows the sigma-radius curves on ImageNet. 
Notably, a larger number of data points do not satisfy the concavity and quasiconcavity. 
A plausible explanation for this discrepancy lies in the increased complexity of the data manifold for classifiers operating on ImageNet, which is attributed to its higher number of classes and dimensions. This elevated complexity results in intricate and non-smooth decision boundaries. 
Consequently, employing a Gaussian kernel to smooth the data space can lead to unstable results.
Nonetheless, we underscore the fact that a significantly larger number of data points adhere to quasiconcavity rather than concavity. This observation highlights the advantage of leveraging quasiconcavity for addressing optimization problems within the domain of randomized smoothing.

\section{Comparison with Prior Work}\label{appdix:comparison}
\textbf{Detailed comparisons with DDRS: }
The proposed QCRS method is novel in that it requires a much looser $(\upsilon^-,\upsilon^+)$-SQC assumption and obtains the optimal sigma solution with a much faster convergence rate than prior work. 
We illustrate the detailed comparison to the gradient-based method.
Table~\ref{tab:comp_ddrs} compares QCRS to DDRS \cite{alfarra2020data}, a gradient-based SOTA method for finding the optimal sigma.
\begin{table}[t]
    \centering
    \caption{Detailed comparisons with DDRS.}
    \begin{tabular}{l|c|c}
    \toprule
      & QCRS & DDRS\\
      \midrule
     Assumption & $(\upsilon^-,\upsilon^+)$-SQC & concavity \\
     Generality  & 	high (99\%)	 & low (7\%) \\ 
     Convergence & $\mathcal{O}((\frac{1}{2})^t)$ & no \\
     Back-propagation & no & yes \\
     Time cost    & low & high\\
     \bottomrule
    \end{tabular}
    \label{tab:comp_ddrs}
\end{table}
For the generality of assumptions, almost all data points satisfy the $(\upsilon^-,\upsilon^+)$-SQC condition, while only $6.7\%$ of data points satisfy concavity. 
In addition, the proposed QCRS guarantees convergence, and the complexity is $\log N$ to find the optimal solution on the $N$-grid.
Furthermore, QCRS enjoys the advantages of computation. 
It does not employ back-propagation, making it computationally less expensive than DDRS.

\noindent\textbf{Relative comparisons with prior work: }
When reproducing the results of prior work, it is important to consider that randomized smoothing involves random components, such as Monte Carlo sampling, and different studies may have slight variations in parameter selection. 
These factors can contribute to variations in the obtained results, even when attempting to reproduce the same experiments. 
To provide a comprehensive comparison, in addition to reporting our reproduced DDRS and DSRS results, we also include the original results from the DDRS and DSRS papers as a reference. 
Since these papers commonly use the original \textsc{Cohen} method as a baseline, we can measure the improvement achieved by DDRS/DSRS compared to the \textsc{Cohen} baseline in their respective papers.
This improvement is referred to as the relative improvement.
Table~\ref{tab:dsrs1} presents the relative improvement in certified accuracies under different radii for DSRS and DDRS, with the values reported directly from the original DDRS/DSRS papers.

\begin{table*}[t]
    \centering
    \caption{Certified accuracy under different radii $r$ of DSRS, DDRS, Grid Search, and the proposed QCRS. The ``Cohen'' values \cite{li2022double,alfarra2020data} of DSRS/DDRS are reported from their original paper. ``+$\%$'' indicates the relative improvement compared to the respective \textsc{Cohen} baseline.}
    \vspace{-0.1cm}
    \scalebox{0.86}{
    \begin{tabular}{l|ccccccccc}
    \toprule
         & \multicolumn{9}{c}{Certified Accuracy}\\
        Certified radii $r$ & $0.25$& $0.5$& $0.75$& $1.0$& $1.25$& $1.5$& $1.75$& $2.0$& $2.25$\\
        \midrule
        \midrule
        \cohen \cite{li2022double} & 0.56 & 0.41 & 0.28 & 0.19 & 0.15 & 0.10 & 0.08 & 0.04 & 0.02\\
        ~+DSRS \cite{li2022double} & 0.57 & 0.43 & 0.31 & 0.21 & 0.16 & 0.13 & 0.08 & 0.06 & 0.04\\
        ~~~(+\%) & 1.8\% & 4.9\% & 10.7\% & 10.5\% & 6.7\% & 30.0\% & 0.0\% & 50\% & 100\%\\
        \midrule
        \cohen \cite{alfarra2020data} & 0.58 & 0.40 & 0.29 & 0.20 & 0.13 & 0.07 & 0.03 & 0.00 & 0.00\\
        ~+DDRS \cite{alfarra2020data} & 0.65 & 0.48 & 0.38 & 0.28 & 0.17 & 0.08 & 0.03 & 0.01 & 0.00\\
        ~~~(+\%) & 12.1\% & 20.0\% & 31.0\% & \textbf{40.0\%} & 30.8\% & 14.3\% & 0.0\% & NA & 0.0\% \\
        \midrule
        \cohen (Ours) & 0.55 & 0.41 & 0.32 & 0.23 & 0.15 & 0.09 & 0.05 & 0.00 & 0.00\\
        ~+Grid Search (24 points) & 0.58 & 0.51 & 0.42 & 0.30 & 0.18 & 0.12 & 0.07 & 0.04 & 0.01\\
        ~~~(+\%) & 5.5\% & 24.4\% & 31.2\% & 30.4\% & 20.0\% & \textbf{33.3\%} & \textbf{40\%} & NA & NA \\
        \midrule
        \cohen (Ours) & 0.55 & 0.41 & 0.32 & 0.23 & 0.15 & 0.09 & 0.05 & 0.00 & 0.00\\
        ~+\textbf{QCRS (Proposed)} & 0.64 & 0.54 & 0.43 & 0.31 & 0.20 & 0.11 & 0.05 & 0.02 & 0.01\\
        ~~~\textbf{(+\%)} & \textbf{16.4\%} & \textbf{31.7\%} & \textbf{34.4\%} & 34.8\% & \textbf{33.3\%} & 22.2\% & 0.0\% & NA & NA \\
    \bottomrule
    \end{tabular}}
    \label{tab:dsrs1}
\end{table*}

\noindent\textbf{Experiments on DSRS: }
DSRS \cite{li2022double} uses double-sampling with two different generalized Gaussian distributions to estimate the probability of class A.
The sigma values of those generalized Gaussian distributions are $\sigma=0.5$ and $\sigma=0.4$. 
It is important to note that their models are trained using the generalized Gaussian, which is different from Cohen's method.
To evaluate DSRS, we compare two trained models and three different sampling methods.
First, we compare the original Cohen's model trained by Gaussian noise (denoted as ``G-'') and the DSRS model trained by generalized Gaussian noise according to DSRS (denoted as ``GG-'').
Second, for randomized smoothing sampling, we compare three methods:
\begin{enumerate}
    \item ``-G'' indicates the original Gaussian smoothing with a Monte Carlo sampling number of 100,000.
    \item ``-GG'' indicates the generalized Gaussian smoothing with a Monte Carlo sampling number of 100,000.
    \item ``-DS'' indicates the double-sampling generalized Gaussian smoothing. One is with $\sigma=0.5$, and a Monte Carlo sampling number of 50,000, and the other is with $\sigma=0.4$ and a Monte Carlo sampling number of 50,000.
\end{enumerate}

\begin{figure}[t]
    \centering
    \includegraphics[width=0.49\textwidth]{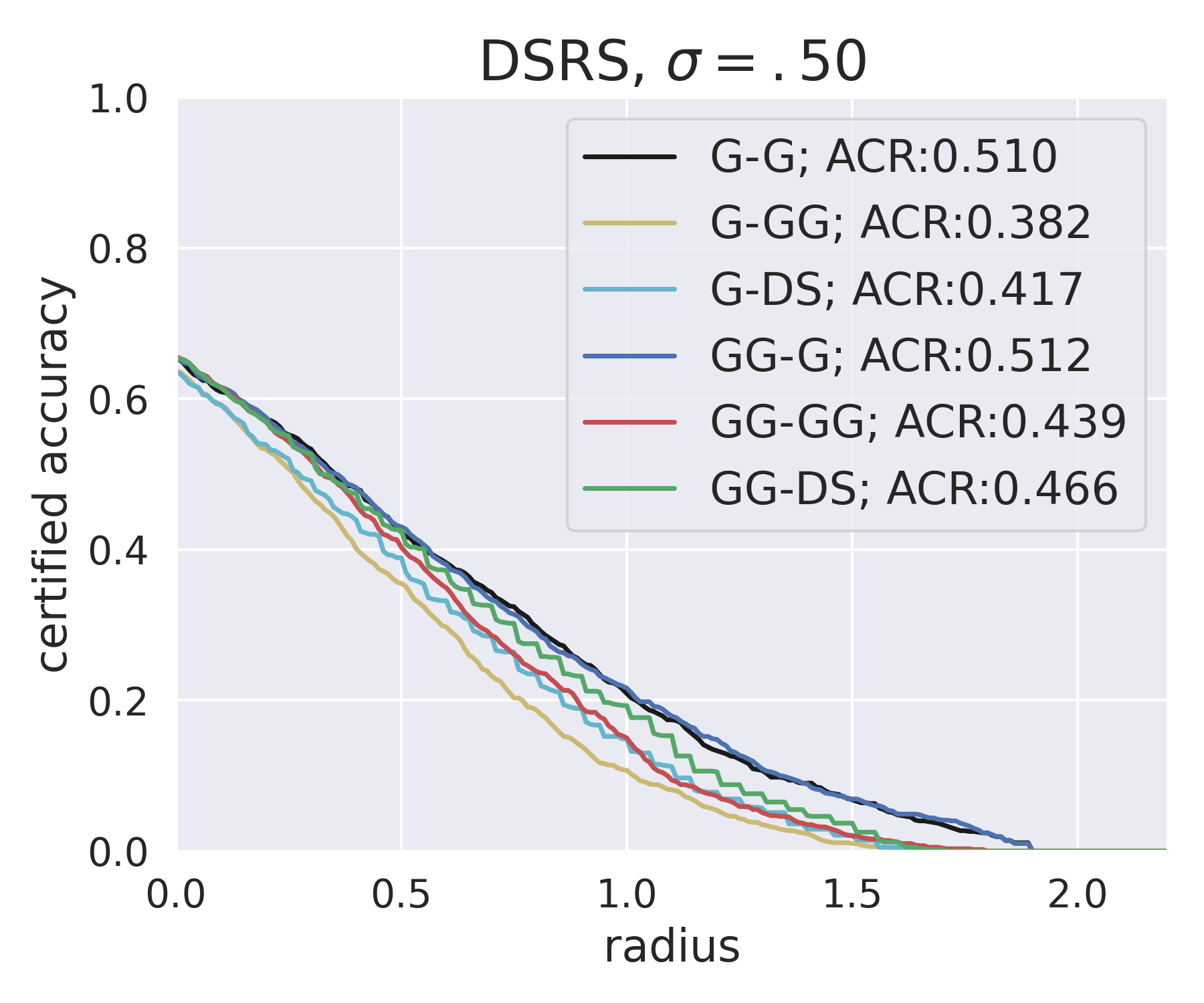}
    \caption{ACR results of DSRS with different settings. The experiments are labeled according to Table~\ref{tab:dsrs_diff}. }
    \label{fig:dsrs_exp}
\end{figure}

Fig~\ref{fig:dsrs_exp} and Table~\ref{tab:dsrs_diff} show the radius-accuracy curves and ACR results, respectively.
However, the results of the double-sampling method did not align with our expectations, despite following the suggestions and source code provided in the DSRS paper. 
We suspect that certain hyperparameters may not have been optimized properly in our experiments. 
In the main paper, we report the last row of Table~\ref{tab:dsrs_diff}, as the hyperparameters used in that row align with the recommendations provided by DSRS.
In addition, we also report the original results in the DSRS paper, as Table~\ref{tab:dsrs1} shows.

\begin{table}[t]
    \centering
    \caption{The ACR results of DSRS on CIFAR-10.}
    \scalebox{0.84}{
    \begin{tabular}{lllc}
    \toprule
        Exp. & Training & Sampling & ACR \\
        \midrule
        G-G & Gaussian & Gaussian &  0.510\\
        G-GG & Gaussian & Generalized Gaussian & 0.382 \\
        G-DS & Gaussian & Double & 0.417 \\
        GG-G & Generalized Gaussian & Gaussian & 0.512 \\
        GG-GG & Generalized Gaussian & Generalized Gaussian & 0.439 \\
        GG-DS & Generalized Gaussian & Double & \textbf{0.466} \\
        \bottomrule
    \end{tabular}}
    \label{tab:dsrs_diff}
\end{table}

\section{Proofs}\label{appendix:proof}

\subsection{Proof of Lemma 1}
We first prove Lemma 1 of the main paper.
\begin{proof}
Suppose $h$ is a strictly quasiconcave function and $x^*$ is a local maximum of $h$. We want to show that $x^*$ is also a global maximum of $h$.
Assume, for the sake of contradiction, that $x^*$ is not a global maximum. This means that there exists a point $y$ in the domain of $h$ such that $h(y) > h(x^*)$.
Let's consider the point $z = \lambda x^* + (1-\lambda)y$, where $\lambda$ is chosen such that $0 < \lambda < 1$. Since $x^*$ is a local maximum, there exists a neighborhood around $x^*$ where $h(x^*) > h(z)$.
Now, by the definition of strict quasiconcavity, we have:
\[
h(z) > \min\{h(x^*), h(y)\}.
\]
Since $h(y) > h(x^*)$, we have $h(z) > h(x^*)$.
This contradicts the assumption that there exists a point $y$ with $h(y) > h(x^*)$. Therefore, the assumption that $x^*$ is not a global maximum must be false.
Hence, we conclude that if $h$ is a strictly quasiconcave function and $x^*$ is a local maximum of $h$, then $x^*$ is also a global maximum of $h$.
\end{proof}

\subsection{Proof of Lemma 2}
We begin by proving the ``$\Rightarrow$'' direction, followed by the ``$\Leftarrow$'' direction.
Proving Lemma 2 simplifies when we apply the first-order condition of quasiconcavity as outlined in \cite{boyd2004convex}.
The first-order condition of quasiconcavity is as follows:
\begin{proposition}
\label{prop:first_order}
     (First-order condition \cite{boyd2004convex}) Suppose $h: \mathbb{R}^n \rightarrow \mathbb{R}$ is differentiable. Then $h$ is strictly quasiconcave if and only if
    \[
    h(y)>h(x) \Rightarrow \nabla h(x)(y-x) > 0.
    \]
\end{proposition}
First, we employ Proposition~\ref{prop:first_order} to prove the ``$\Rightarrow$'' direction of  Lemma 2.
\begin{proof}
Suppose $h$ is a strictly quasiconcave function and $x^*$ is a local maximum of $h$.
For any other points $x$, it holds that $h(x^*)>h(x)$.
Consequently, when $x < x^*$, the inequality $\nabla h(x)^T(x^*-x) > 0$ ensures that $\nabla h(x)>0$.
Conversely, if $x > x^*$, the inequality $\nabla h(x)^T(x^*-x) > 0$ ensures that $\nabla h(x)<0$.
That is, $\nabla h(x)$ positive on the left side of the optimal solution and negative on the right side.
\end{proof}
Next, we employ Proposition~\ref{prop:first_order} to prove the ``$\Leftarrow$'' direction of Lemma 2.
\begin{proof}
Consider a function $h$ with a unique optimal solution $x^*$. Furthermore, $h$ satisfies the condition that $\nabla h(z)>0$ for $z<x^*$ and $\nabla h(z)<0$ for $z>x^*$.
Consider two points $x$ and $y$. If $x<x^*$, then for any $y$ such that $h(y)>h(x)$, it follows that $y$ must be greater than $x$ because $\nabla h(z)>0$ for $z<x$. Therefore, $\nabla h(x)(y-x)>0$.
On the other hand, If $x>x^*$, for any $y$ such that $h(y)>h(x)$, it follows that $y$ must be smaller than $x$ because $\nabla h(z)<0$ for $z>x$. Therefore, $\nabla h(x)(y-x)>0$.
Thus, according to Proposition~\ref{prop:first_order}, $h$ is a strictly quasiconcave function.
\end{proof}


\subsection{Proofs of convergence analysis}
In this section, we theoretically analyze the convergence of the proposed methods and provide proof.
The convergence of the gradient-based method is also discussed.

Without loss of generality, we discuss convexity rather than concavity here.
First, we prove Theorem~1 in the main paper as follows:
\begin{proof}
Let $\sigma_t$ be the $\sigma$ under $t$ iterations.
Suppose that $R$ satisfies $(\upsilon^-,\upsilon^+)$-SQC condition with $(\upsilon^-,\upsilon^+)=(1,1)$, and there exists a $\sigma^* \in [\sigma_{min}, \sigma_{max}]$. Then, for the first iteration $\sigma_1 = \frac{ \sigma_{max} + \sigma_{min}}{2}$, we have
\begin{align*}
    \frac{ \sigma_{max} - \sigma_{min}}{2} \geq | \sigma_1 - \sigma^*|,
\end{align*}
because $\sigma_1$ is the midpoint of $\sigma_{min}$ and $\sigma_{max}$.
Without loss of generality, we assume $\sigma_{min} \leq \sigma^* \leq \sigma_1$. Thus, according to Algorithm~1, $\sigma_2 = \frac{\sigma_{min}+\sigma_1}{2}$, and 
\begin{align*}
    \frac{ \sigma_{max} - \sigma_{min}}{2^2} \geq | \sigma_2 - \sigma^*|.
\end{align*}
If we run $t$ iteration, we can conclude that
\begin{align*}
    \frac{ \sigma_{max} - \sigma_{min}}{2^t} \geq | \sigma_t - \sigma^*|.
\end{align*}
\hfill$\blacksquare$
\end{proof}

Therefore, to achieve $\delta$-optimal, the convergence rate of the proposed method is $\mathcal{O}((\frac{1}{2})^t)$.

Next, we further analyze the convergence of the gradient-based method.
$L$-smoothness or $\mu$-strong convexity are necessary to guarantee convergence.
We first show the convergence rate of $L$-smoothness:
\begin{proposition}\label{thm:lsmooth}
Suppose a function $R(\sigma)$ is $L$-smooth for some $L > 0$ with respect to $\sigma$. Then, if we run gradient descent for $t$ iterations, it converges as follows \cite{nesterov2018lectures}:
\begin{align*}
R(\sigma_t) - R(\sigma^{*}) \leq \frac{L|\sigma_1 - \sigma^{*}|^2}{2(t-1)}.
\end{align*}
\end{proposition}

In addition, if both $L$-smoothness and $\mu$-strong convexity hold, the convergence rate is as follows:
\begin{proposition}\label{thm:muconvex}
Suppose a function $R(\sigma)$ is $L$-smooth and $\mu$-strongly convex for some $L, \mu >0$ with respect to $\sigma$, and $\hat{\sigma}$ is the optimal sigma. 
Then, if we run gradient descent for $t$ iterations, it converges as follows \cite{nesterov2018lectures}:
\begin{align*}
|\sigma_t - \sigma^{*}|^2 \leq (\frac{L-\mu}{L+\mu})^{(t-1)
}|\sigma_1 - \sigma^{*}|^2.
\end{align*}
\end{proposition}

Proposition~\ref{thm:lsmooth} shows the convergence rate under the convex and $L$-smooth condition. 
$R$ with $L$-smoothness can not guarantee $\delta$-optimal, i.e., $|\sigma^* - \sigma|\leq \delta$.
On the other hand, Proposition~\ref{thm:muconvex} shows the convergence rate under the $L$-smooth and $\mu$-convex condition, which is faster but stricter than Proposition~\ref{thm:lsmooth}.
If we want to achieve $\delta$-optimal for $\sigma$, the convergence rate of $\mathcal{O}((\frac{L-\mu}{L+\mu})^t)$, where $t$ is the number of iterations.

\section{Implementation Details}
We use an NVIDIA GeForce® RTX 3090 GPU and an AMD Ryzen 5 5600X CPU with 32GB DRAM to run the experiments. 
Following prior work, we use ResNet110 for CIFAR-10 and ResNet50 for ImageNet.
We use $500$ as the MC sampling number to estimate gradients in Algorithm~1.
The terminal condition of Algorithm~1 depends on $\varepsilon$, which indicates an $\varepsilon$-optimal solution, i.e., $|\sigma - \sigma^*|\leq \varepsilon$.
The $\varepsilon$ we used in our experiments is $0.01$, and $\tau$ (the step to compute gradient) is $\pm 0.05$.
Regarding grid search, we use 24 points for CIFAR-10 and 11 points for ImageNet.
The searching region is $0.08$ to $0.50$ for $\sigma=0.12$, $0.15$ to $0.7$ for $\sigma=0.25$, and $0.25$ to $1.0$ for $\sigma=0.50$. 


\section{Ablation Study}

A comprehensive ablation study for hyperparameter tuning is shown in Table~\ref{tab:ablation_s5}. The terms ``left" and ``right" denote the endpoints $\sigma_{min}$ and $\sigma_{max}$ of the search region. That is, Algorithm 1 would search the optimal sigma value within $[\sigma_{min}, \sigma_{max}]$ interval. Notably, the results illustrate that a larger search region leads to a notable enhancement in ACR. On the other hand, the choice of step distance for gradient calculation is important. Smaller values introduce imprecise gradient directions, while excessively larger values struggle to capture local information effectively. 
These experimental evaluations were conducted on the test set of CIFAR-10, specifically following the condition $id\%50==0$.

\begin{table}[tp]
    \centering
    \begin{tabular}{cccc|c}
    \toprule
         sigma & left ($\sigma_{min}$) & right ($\sigma_{max}$) & step ($\tau$) & ACR\\
         \midrule
         0.5 & 0.12 & 0.8 & 0.4 & 0.619 \\
         0.5 & 0.12 & 1.2 & 0.4 & 0.627 \\
         0.5 & 0.12 & 1.9 & 0.4 & 0.638 \\
         0.5 & 0.15 & 0.9 & 0.4 & 0.624 \\
         0.5 & 0.20 & 0.9 & 0.4 & 0.624 \\
         0.5 & 0.20 & 1.0 & 0.4 & 0.625 \\
         0.5 & 0.20 & 1.2 & 0.4 & 0.628 \\
         0.5 & 0.20 & 1.5 & 0.4 & 0.629 \\
         0.5 & 0.20 & 1.7 & 0.4 & 0.639 \\
         0.5 & 0.30 & 1.0 & 0.4 & 0.622 \\
         0.5 & 0.30 & 1.5 & 0.4 & 0.629 \\
         0.5 & 0.3 & 1.2 & 0.2 & 0.640 \\
         0.5 & 0.3 & 1.2 & 0.4 & 0.634 \\
         0.5 & 0.3 & 1.2 & 0.6 & 0.630 \\
         0.5 & 0.3 & 1.2 & 0.8 & 0.627 \\
         0.5 & 0.3 & 1.2 & 1.2 & 0.615 \\
         0.5 & 0.12 & 1.7 & 0.05 & 0.572 \\
         0.5 & 0.12 & 1.7 & 0.1 & 0.640 \\
         0.5 & 0.12 & 1.7 & 0.2 &\textbf{ 0.644} \\
         0.5 & 0.12 & 1.7 & 0.4 & 0.639 \\
          \bottomrule
    \end{tabular}
    \caption{Hyperparameter tuning ablation study for QCRS. The terms ``left" and ``right" indicate the bounds of the search region. Evidently, broader search regions yield improved ACR. On the other hand, the step distance for gradient computation also plays a crucial role. Smaller values result in less accurate gradient directions, while larger values struggle to capture local information effectively. These experiments were conducted on CIFAR-10 with the condition $id\%50==0$.}
    \label{tab:ablation_s5}
\end{table}

\section{Error on Sigma}

\begin{figure*}[t]
\centering
\includegraphics[width=0.75\linewidth]{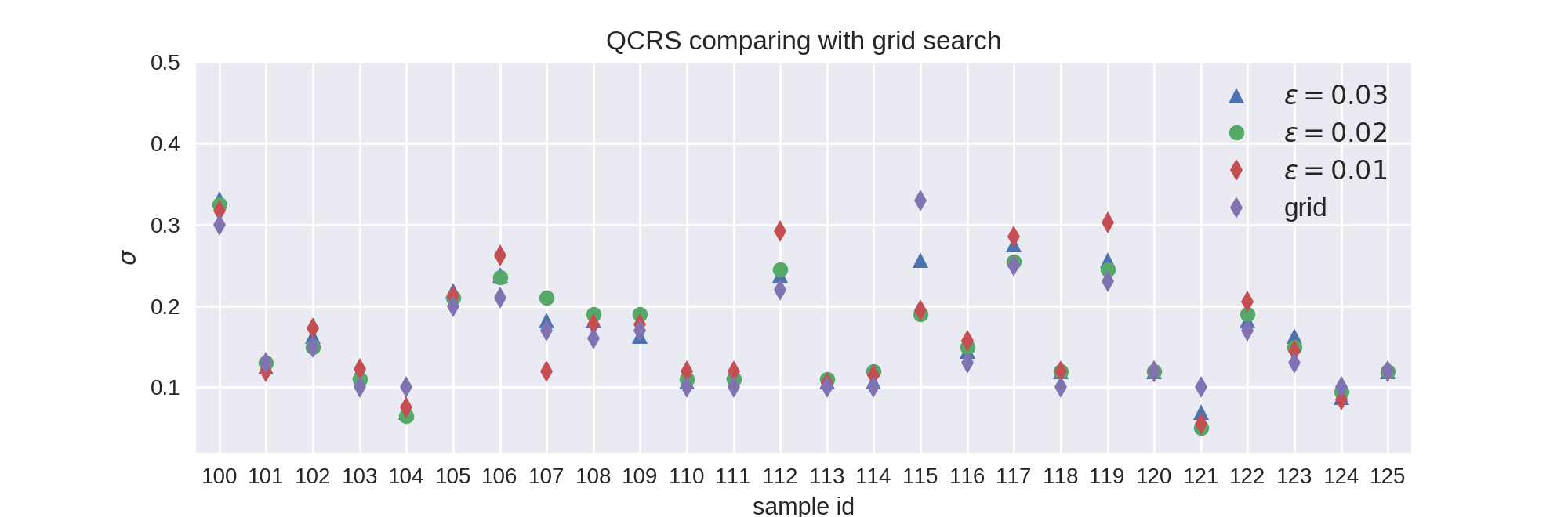}%
\caption{We demonstrate four sigma values found by different algorithms and parameters. The purple points represent sigma values found by grid search. The other points are sigma values found by the proposed QCRS. We test different $\varepsilon$ values, which indicate $\varepsilon$-optimal solutions. The $\varepsilon$ values also serve as the terminal conditions of Algorithm~1.}
\label{fig:sigma_error}
\end{figure*}

We assess the accuracy of the sigma values obtained by QCRS by comparing them to those obtained through grid search. Assuming that the optimal sigma value obtained through grid search represents the true optimum, we evaluate the effectiveness of QCRS in finding the optimal sigma. 
We randomly selected several data points for evaluation, and the results are illustrated in Fig.~\ref{fig:sigma_error}. 
It can be observed that the sigma values found by QCRS are in close proximity to the sigma values obtained through grid search. This suggests that QCRS is capable of effectively approximating the optimal sigma value.
Furthermore, we conduct tests on the $\varepsilon$ value in Algorithm 1, which serves as the terminal condition. Using a smaller $\varepsilon$ value can lead to sigma values closer to the optimum found through grid search. However, even with different $\varepsilon$ values, the sigma values obtained by QCRS and grid search remain quite close.
This demonstrates the ability of QCRS to optimize the sigma value for randomized smoothing effectively.

\section{Gradient Stability}
\begin{figure}
    \centering
    \includegraphics[width=0.45\linewidth]{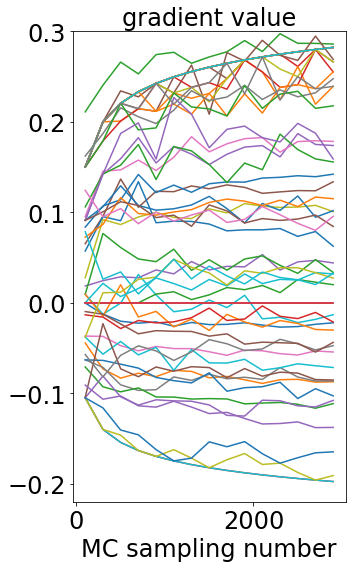}
    \caption{Gradient values with respect to different MC sampling numbers. The different curves indicate different data points. The gradient values with regard to different MC sampling numbers are unstable, but their signs remain the same in almost all cases. The proposed QCRS is stable in the optimization as it only relies on the signs of gradient values.}
    \label{fig:grad_stability2}
\end{figure}

As mentioned in the main paper, using the radius calculated by Monte Carlo sampling as the objective function to optimize the sigma value can result in unstable gradients.
To verify this, we analyzed the gradient values under different MC sampling numbers, as shown in Fig~\ref{fig:grad_stability2}. 
The different curves indicate the different data points.
We found that the gradient values exhibit significant variations when using low MC sampling numbers. 
Consequently, using a low number of Monte Carlo (MC) samples may lead to inaccurate gradient estimates, which can adversely impact gradient-based optimization methods. 
In contrast, the proposed QCRS method solely relies on the sign of the gradient, which is considerably more stable than the actual gradient values.
This can be observed in Fig.~\ref{fig:grad_stability2}, where the gradient sign remains unchanged while the gradient values exhibit instability.

\begin{figure}[t]
\centering
\subfloat[MC sampling]{\includegraphics[width=0.5\linewidth]{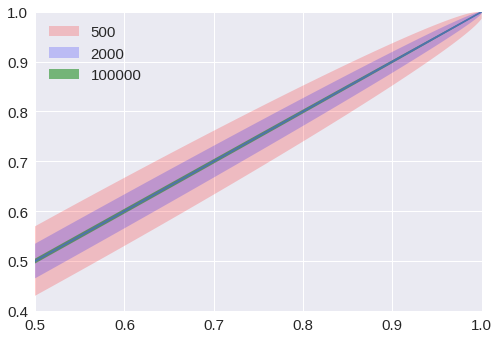}%
}
\subfloat[Zoom in]{\includegraphics[width=0.5\linewidth]{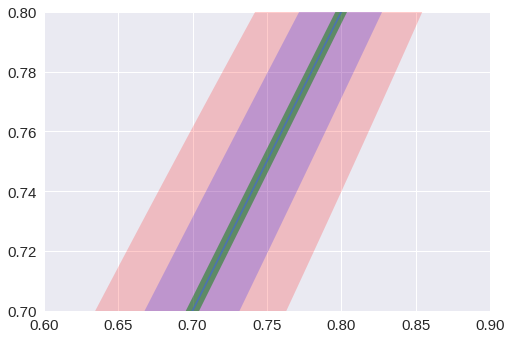}%
}
\caption{We plot the confidence interval of $\pabar$ with respect to $\pabar$. The number of MC sampling significantly affects the confidence interval. Lower MC sample numbers result in wider confidence intervals. The red region in the figure represents the confidence interval of $\pabar$ with a $1-\alpha$ confidence level and $500$ MC sampling number. It is wider than the one with a $100,000$ MC sampling number. This shows that the low MC sampling number makes the certified region uncertain. When estimating gradient values, if the value depends on MC sampling, the corresponding gradient will be unstable, making optimization difficult to converge. Prior work, such as DDRS, used a very low MC sampling number (as low as eight) when optimizing the sigma value, leading to unstable optimization.}
\label{fig:mc_demo}
\end{figure}

In addition, as mentioned in the main paper, the MC sampling number significantly affects the estimation of $\pabar$. 
As shown in Fig.~\ref{fig:mc_demo}, when the MC sampling number is $500$, the possible interval is represented by the red region with a confidence level of $1-\alpha$. 
The wide extent of the red region indicates a high level of uncertainty in the estimation of $\pabar$, resulting in an unstable estimate. 
Previous work often employs low MC sampling numbers to minimize computational costs in the context of backpropagation, which further contributed to unstable computed gradients and inaccurate $\pabar$ estimation. These factors can ultimately lead to poor optimization of $\sigma$.

\section{Limitations}
\textbf{Generality of quasiconcavity: }
The theoretical generality of the quasiconcavity assumption for data points remains unknown. While we empirically demonstrate the presence of quasiconcave sigma-radius curves in most data points in CIFAR-10 and ImageNet, a rigorous theoretical proof is lacking. A further theoretical investigation is required to determine the extent to which the quasiconcavity property holds across different datasets and problem domains. 

\noindent\textbf{Convergence assumption: }
The convergence guarantee of Algorithm 1 relies on the $(\upsilon^-,\upsilon^+)$-SQC condition. 
If this condition fails for a specific data point, the proposed QCRS algorithm may not be able to find the optimal sigma value. 
Therefore, the effectiveness of the algorithm is contingent on the validity of this assumption.

\noindent\textbf{Computational cost: }
While the proposed algorithm aims to be efficient, it still introduces additional computational overhead compared to traditional randomized smoothing techniques. 
Furthermore, randomized smoothing itself is computationally expensive. As a result, the overall computational cost of the proposed method may limit its applicability in resource-constrained environments or real-time applications. 
Additionally, tuning the hyperparameters in QCRS may require additional computational resources and experimentation. 
Therefore, it is important to carefully consider the computational requirements and practical feasibility of the proposed method in real-world scenarios.

\section{Constant Sigma}
A consistent sigma may be required during deployment, presenting a limitation and challenge for input-specific RS. 
The memory bank approach suggested by DDRS is a simple and effective solution and is compatible with our proposed method. 
With regard to the proposed QCRS, employing a constant sigma during deployment is indeed feasible, though its applicability is confined to particular scenarios.
For example, consider binary certified robust classification. \footnote{The certification experiment in Cohen's paper also employs binary settings. They use $p_B = 1-p_A$.}
In such cases, our approach can transition from datapoint-wise perspective to a broader class-wise perspective. 
Before deployment, we utilize QCRS to estimate an optimized sigma tailored to a specific class, say ``cat''. 
At deployment, we consistently employ this optimized constant sigma value to certify and predict ``cat'' and ``non-cat''. 
Table~\ref{tab:classwise} illustrates the class-wise ACR of this scenario on CIFAR-10. 
Each class in Table~\ref{tab:classwise} has a tailored sigma value (the values come from Table 5 in the main paper), and this class-wise constant sigma value is used to certify and predict for the binarized task.
This approach substantially increases the certified radii by approximately $40\%$ for classes with smaller radii, including cats, birds, and deer.
In CIFAR-10, only one class, cars, does not show equally favorable results. This is because ``cars'' already have an adequate radius with the standard sigma; a larger sigma may lead to misclassifications and zero radii for certain ``car'' instances.

Furthermore, to illustrate how to use constant sigma at deployment, we take the binary face recognition task in unlocking mobile phones as a practical example.
When vendors distribute the model weights to individual users, we employ the proposed method to compute the new sigma value tailored to each user. 
This customization allows us to seamlessly construct a more robust smoothed classifier.
In this scenario, every user has their own optimized sigma value, which remains constant.
By introducing the sigma value as a new parameter, we can easily improve the certified radii for the system via a single parameter instead of all parameters in the base classifier. 
That is, the proposed method incurs negligible additional costs and provides larger radii at deployment compared to the other methods, such as grid search and retraining the base classifier. 
Given the increasing complexity of DNN models and the impracticality of retraining or fine-tuning, our low-cost QCRS emerges as a practical and efficient alternative.

\begin{table}[t]
    \centering
    \begin{tabular}{lccccc}
    \toprule
       ACR  &  birds & cats & deer &  trucks \\
       \midrule
       $\sigma=0.5$  & .317 & .250 & .217 & .705 \\
       Class-wise $\sigma$ & \textbf{.457} & \textbf{.337}  & \textbf{.302 } & \textbf{.788} \\
    \bottomrule
    \end{tabular}
    \caption{Class-wise QCRS. We use the constant sigma in Table 5 of the main paper for different classes throughout the binary tasks.}
    \label{tab:classwise}
\end{table}


\end{document}